\documentclass[11pt,a4paper]{article}

\usepackage{authblk}
\usepackage{amsmath,amssymb}
\usepackage{bm}
\usepackage{tikz}
\usetikzlibrary{arrows.meta, positioning,shapes.geometric}
\usepackage{fancyvrb}
\usepackage{fvextra}
\usepackage{subcaption}

\usepackage{booktabs}
\usepackage{tabularx}
\usepackage{array}
\usepackage{longtable}
\usepackage{pdflscape}

\title{
Depth-Wise Inference Dynamics in Transformers:
An Intervention-Based Decomposition of Self-Attention and MLP Contributions
}
\author[]{Kiyoshi NISHIKAWA}
\author[]{Yukihisa SUZUKI}
\author[]{Norio TAGAWA}
\affil[]{Department of Systems Design, Tokyo Metropolitan University \\ \texttt{\{kiyoshi, y\_suzuki, tagawa\}@tmu.ac.jp}}

\date{\today}

\begin{document}
\maketitle

\begin{abstract}
Transformer-based models achieve strong performance across many machine-learning tasks, yet the computation they perform across depth at inference time still lacks a sufficiently explicit dynamical characterization at the level of distinct residual components.
In this work, we study Transformer inference across depth by formulating its layer-wise computation as a discrete-time state-update process on the full token sequence, with each residual update decomposed into additive Self-Attention and MLP contributions. 
This decomposition enables component-wise interventions during inference and provides an operational framework for probing internal computation through intervention and response.
Within this framework, we introduce output-based operational notions of invariance region and basin-like selection on the sequence-level state space, and we study how these notions evolve across depth under selective suppression of attention-side updates.
Our experiments on a trained GPT-style model suggest that, for later intervention
onset depths, suppressing the context-side attention-output contribution generally
produces smaller changes in the measured position-wise prediction diagnostics, while
the remaining depth-wise dynamics continue in a token-dependent and non-uniform manner.
This supports a structured view in which Self-Attention primarily mediates sequence-level reorganization and context integration, whereas the MLP mainly supports token-wise local refinement and stabilization within an already selected output-associated region.

The contribution of this work is not to claim that internal Transformer
dynamics can be quantitatively probed in general, but to organize such probing
around intervention-based output-side stability at token-level resolution.
In this sense, the proposed framework provides a minimal and measurable description of Transformer inference as an intervenable dynamical process and offers a dynamical interpretation of how depth-wise computation separates context mixing from subsequent local refinement.
\end{abstract}

\section{Introduction}

Transformer architectures are central to modern machine learning and underpin strong performance across natural language processing, vision, and multimodal modeling \cite{vaswani2017attention,dosovitskiy2021vit}. Their computation is built from repeated compositions of Self-Attention, position-wise nonlinear transformations (MLPs), and residual connections. Because of this residual structure, inference across depth can naturally be viewed as an iterative state-update process and, in suitable abstractions, as a dynamical system evolving over layers \cite{lu2019mpds,geshkovski2023clusters,geshkovski2023mathperspective}. At the same time, a precise characterization of what is computed across depth, and which component of the residual update is responsible for which aspect of that computation, remains incomplete.

A substantial body of prior work has already shown that internal computation in Transformer language models can be probed quantitatively. Examples include readout-based methods such as the logit lens \cite{nostalgebraist2020logitlens} and its refinement via the tuned lens \cite{belrose2023tunedlens}, probing and representation-analysis approaches more broadly \cite{belinkov2022analysis}, and intervention-based mechanistic studies such as causal tracing, model editing, and related activation-patching-style analyses \cite{meng2022rome}. In parallel, recent studies have reported that the functional contribution of Transformer sublayers is strongly non-uniform across depth: interventions on upper attention layers can have surprisingly small effects in some regimes, whereas MLP-side perturbations often remain much more consequential \cite{benartzy2024attendfirst,he2024notallattention,lad2024stages}. These findings make it increasingly implausible to describe Transformer inference as a homogeneous sequence of equally necessary layer transformations.

However, several important gaps remain. First, many analyses are evaluated at the level of aggregate task accuracy, averaged prompts, or layer-wise summaries, which can obscure token-to-token heterogeneity within a single forward pass. Second, even when upper-layer attention redundancy is observed, the resulting computation is not always organized into an explicit dynamical interpretation that distinguishes \emph{context-side mixing}, \emph{output-side readout}, and \emph{post-selection local refinement}. Third, while dynamical-systems viewpoints on Transformers have been developed \cite{lu2019mpds,geshkovski2023clusters,geshkovski2023mathperspective}, they are not always paired with a concrete intervention-based operational criterion for identifying when a certain phase of computation has effectively completed.

The present work addresses these gaps by decomposing each residual update into an attention contribution and an MLP contribution, and by introducing inference-time interventions that selectively suppress one contribution beyond a chosen depth. This allows us to probe Transformer computation not only as a stack of layers, but as a structured dynamical process composed of distinguishable residual components. In particular, we distinguish intervention regimes that suppress \emph{context-side attention mixing} while preserving the output-token readout channel from those that suppress attention-side contribution more broadly. These interventions make it possible to ask, in an operational sense, at which depth the output for an individual token becomes effectively invariant to further context-side attention mixing.

This shift in viewpoint is central to the contribution of the paper. 
Our claim is not that internal states of Transformers can be observed
quantitatively for the first time; rather, our contribution is to organize such
observation around intervention-response analysis of token-level output-side
stability.
Concretely, for each token, we examine how the measured intervention-response
differences change with the intervention onset depth, yielding a qualitative
characterization of token-wise stabilization behavior. This reveals that depth-wise
computation is markedly non-uniform across tokens even within the same prompt.
In this way, we move from a purely layer-wise or task-aggregate description toward an operational description of trajectory-level inference dynamics. In this sense, the present work shifts the focus from detecting internal signals to structurally organizing their depth-wise evolution under controlled interventions.

This token-resolved intervention framework also supports a dynamical interpretation. We describe the inference process in terms of a depth-indexed progression on the sequence-level state space, where interventions help identify a transition between two qualitatively different regimes. In the earlier regime, Self-Attention plays the primary role in sequence-level reorganization and context integration. In the later regime, the dominant remaining changes are more consistent with token-wise local refinement and stabilization, for which the MLP contribution is comparatively more important. This interpretation is also consistent with prior analyses that emphasize distinct functional roles for feed-forward layers, including their interpretation as key-value memories and as additive concept-promoting updates in vocabulary space \cite{geva2021ffnkv,geva2022ffnconcepts}, as well as with theoretical work suggesting a separation between information selection and retention \cite{xu2025filteringstoring}.

Our perspective is also related to adaptive-depth and early-exit research, while differing from it in emphasis. Prior work has shown that different tokens often require different amounts of computation, and that adaptive-depth mechanisms can exploit this heterogeneity by allocating computation dynamically across tokens \cite{elbayad2020depthadaptive,schuster2022calm,jiang2024dllm,luo2025flexidepth,jaber2026tide}. We share the observation of token-wise non-uniformity, but our aim is not primarily inference acceleration. Rather, we interpret token-wise depth heterogeneity as evidence for a structured division of labor between attention-mediated context mixing and subsequent local refinement. Our analysis therefore complements adaptive-depth research by providing an intervention-based operational characterization of token-wise effective depth, rather than directly optimizing computation allocation.

To avoid overstatement, we emphasize that the present notions of invariance region and basin are operational and finite-horizon: they are defined with respect to the remaining depth-wise flow under the trained network, rather than as global attractor basins in the strict autonomous dynamical-systems sense. Within this scope, however, they provide a compact and measurable language for organizing how depth-wise Transformer computation changes under controlled interventions.

The main contributions of this paper are as follows:
\begin{itemize}
    \item We formulate Transformer inference as a discrete-time dynamical process on the sequence-level state space by decomposing each residual update into additive Self-Attention and MLP contributions.
    \item We introduce intervention schemes that selectively suppress residual components during inference and thereby provide an operational method for analyzing internal computation through intervention and response.
      
      \item We define an operational notion of token-level output-side stability under
      intervention and use the resulting intervention-response curves to characterize token-dependent stabilization behavior, revealing non-uniform depth-wise dynamics within individual prompts.

    \item We propose a dynamical interpretation in which attention primarily supports context-side mixing and sequence-level reorganization, whereas the MLP primarily supports local refinement and stabilization after an output-associated region has effectively been selected.
\end{itemize}

The rest of the paper is organized as follows. Section~\ref{sec:related} reviews prior work and clarifies the relation of the present study to interpretability, adaptive computation, and dynamical-systems perspectives. 
Section~\ref{sec:struct_DDS} introduces the formulation, notation, residual decomposition, and output-based criterion for effective invariance.
Section~\ref{sec:DRofSAandMLP} develops the depth-wise interpretation of the roles of Self-Attention and the MLP within the proposed dynamical framework.
Section~\ref{sec:experiment_early_exit} defines the concrete intervention settings and presents the experimental results and token-wise invariance analysis.
Section~\ref{sec:interpretation} discusses the implications, limitations, and broader interpretation of the proposed framework.

\section{Related Work and Positioning}
\label{sec:related}

A growing body of work seeks to understand how Transformers compute, not only in terms of predictive performance, but also in terms of their internal representations, functional organization, and depth-wise processing. In machine learning, this question has been approached from several directions, including layer-wise analysis, mechanistic interpretability, functional decomposition, learning dynamics, adaptive computation, and information-theoretic viewpoints. These lines of work have each provided useful partial characterizations of Transformer computation. However, they do not yet yield a unified framework for describing inference-time depth-wise computation in a form that is both observable and amenable to intervention.

Our starting point is the following gap: most existing analyses describe \emph{what} changes across layers, but they do not directly characterize depth-wise computation as a structured dynamical process whose constituent update components can be separately probed. In particular, when each layer is treated as a single transformation, the contributions of Self-Attention and the MLP remain entangled within the same update, making it difficult to ask operational questions such as when the sequence-level trajectory becomes effectively fixed with respect to the output, or which component continues to matter at a given depth. This section reviews the main relevant lines of work and clarifies how the present study is positioned relative to them.

\subsection{Layer-wise Analysis and Its Limitations}

Layer-wise analysis has been one of the most common approaches for studying the internal behavior of Transformers \cite{rogers2020primer,belinkov2022analysis,clark2019attention,kovaleva2019bert,tenney2019bertflow}. These studies analyze how model behavior changes across depth using attention distributions, representation statistics, probing, and visualization. Related work has also examined redundancy and specialization in attention heads \cite{michel2019heads,voita2019heads}. Such analyses have established that different layers often exhibit distinct functional characteristics, and that both attention patterns and representation statistics change systematically with depth.

A related line of work seeks to read out intermediate model predictions directly from hidden states. Methods such as the logit lens and the tuned lens interpret intermediate residual-stream states as partial latent predictions of the next token, thereby making prediction trajectories across depth quantitatively observable \cite{nostalgebraist2020logitlens,belrose2023tunedlens}. These approaches are highly relevant to the present study because they demonstrate that internal Transformer computation can be probed beyond final-layer outputs. However, they remain primarily readout-based: they do not define output invariance through controlled suppression of decomposed residual contributions, nor do they explicitly separate context-side attention mixing from subsequent local refinement.

Another closely related line of research uses causal intervention to localize and manipulate internal computations. Causal tracing and model-editing studies have shown that specific internal activations and modules can play decisive roles in factual recall and prediction formation \cite{meng2022rome}. These works are important precursors to the present study in that they move beyond passive observation and treat internal perturbation as a tool for functional analysis. Our setting differs, however, in that we do not aim to edit model behavior or localize a single factual association; instead, we study how sustained component-wise suppression across depth reveals token-wise output invariance and effective computational depth.

However, much of this literature still adopts the layer as the minimal unit of analysis. In that setting, Self-Attention and the MLP are aggregated into a single update map, even though they implement qualitatively different operations. As a result, layer-wise analysis provides a useful coarse-grained description of where certain behaviors appear, but it does not by itself provide a component-level account of how those behaviors are produced by the underlying update structure.

\paragraph{Why layer-wise observation alone makes basin-like descriptions difficult.}
This limitation becomes particularly relevant when one seeks to describe Transformer inference in terms of region selection or output invariance on a state space. There are at least three reasons for this. First, when additive terms with different roles remain entangled within a layer, it becomes difficult to determine which component contributes primarily to sequence-level reorganization and which contributes primarily to token-wise local stabilization. Second, statistics averaged over layers, tokens, or inputs can obscure the token-dependent and asynchronous dynamics that coexist within the same layer. Third, to discuss a basin-like region as a set of states associated with the same output, passive observation alone is not sufficient: one needs an operational criterion based on intervention, namely the ability to alter one part of the update and measure whether the output remains invariant.

For this reason, the present work does not reject the layer-wise perspective;
rather, it extends it into a component-wise and intervention-based account
of depth-wise computation.
This makes it possible to study depth-wise computation not only as a sequence of layers, but also as a structured process composed of distinguishable update components.

\paragraph{Observability, identifiability, and intervention.}
More broadly, this viewpoint is consistent with a general principle in the analysis of complex systems: internal structure is often easier to characterize through intervention and response than through passive observation alone. In control and system identification, observability and identifiability formalize the extent to which internal states or parameters can be inferred from external outputs. In large, high-dimensional systems, direct reconstruction of the internal state is often difficult or impossible from observation alone. Motivated by this perspective, we treat selective suppression of an update component not as an architectural modification, but as a measurement operation that reveals how depth-wise computation depends on the constituent parts of the residual update.

\subsection{Functional Decomposition and Interpretability}

Research on the separation of roles between attention and the MLP has also progressed \cite{olah2020zoom,elhage2021framework,xu2025filteringstoring}. For a one-layer Transformer, Xu et al.\ theoretically showed that Self-Attention is involved in the selection and extraction of information, whereas the MLP is involved in the retention of knowledge \cite{xu2025filteringstoring}. Moreover, mechanistic interpretability research has repeatedly suggested that Transformer components behave as distinct functional modules \cite{olah2020zoom,elhage2021framework}. The present work is consistent with these perspectives, but differs in that it emphasizes how this division of labor unfolds dynamically across depth during inference.

\subsection{Learning Dynamics Perspective}

JoMA by Tian et al.\ \cite{tian2024joma} describes the training process of Transformers as a continuous-time dynamical system. By removing Self-Attention and analyzing the resulting process as MLP dynamics, that framework makes a tractable theoretical treatment possible. JoMA also reports temporal structural changes during training, such as early sparsification and later densification of attention. In contrast, the present work focuses on inference rather than training, and addresses how already acquired structure is used across depth. Thus, the learning-dynamics viewpoint is not identical to our setting, but it is related at the broader level of dynamical description.

\subsection{Information Compression View}

From an information-theoretic viewpoint, the computation performed by a Transformer has been described as a three-stage process involving mixing, compression, and refinement \cite{tishby2000information,shwartz2017opening,queipo2025attentionsinks}. These results are broadly compatible with the regime structure suggested in the present work, but the object of analysis is different:
\begin{itemize}
\item Prior work: changes in statistical quantities such as entropy,
\item This work: the structure of the sequence-level state space and the transitions observed on it.
\end{itemize}
Thus, the present study does not deny compression-based interpretations; rather, it seeks to redescribe some of the same phenomena in terms of state-space dynamics.

\subsection{Comparison with Prior Related Methods: Token-Wise Adaptive Depth and Dynamical Interpretation}

Recent work on Transformers and large language models has increasingly explored token-wise adaptive computation and early-exit strategies, motivated by the observation that different tokens require different amounts of computation. Depth-Adaptive Transformer already introduced the idea that different inputs or positions may require different effective depth \cite{elbayad2020depthadaptive}. CALM likewise uses confidence-based adaptive computation during generation \cite{schuster2022calm}. More recent methods make this adaptivity increasingly explicit at the token level. For example, D-LLM introduces a dynamic decision module at each layer and proposes a framework for adaptively allocating computational resources per token \cite{jiang2024dllm}. FlexiDepth adds a plug-in router to a pretrained LLM and dynamically varies the number of layers used per token; it reports that shallow computation often suffices for repetitive or formulaic tokens, whereas deeper computation is needed for high-uncertainty or computation-bearing tokens \cite{luo2025flexidepth}. AdaPonderLM, for recurrent or iterative Transformers, learns token-wise adaptive depth during pretraining in order to allocate more computation to difficult tokens \cite{song2026adaponderlm}. Post-training early-exit systems such as TIDE likewise estimate the depth at which each token's hidden state has effectively converged and improve inference efficiency by choosing different exit depths per token \cite{jaber2026tide}.

These studies share a concern closely related to the token-wise non-uniform dynamics observed in this work: not all tokens require the same depth of computation. Their main emphasis, however, is on inference efficiency and the allocation of computational resources. They do not aim to interpret differences in depth of computation dynamically in terms of a division of labor between sequence-level reorganization by attention and token-wise local refinement by the MLP \cite{elbayad2020depthadaptive,schuster2022calm,jiang2024dllm,luo2025flexidepth,song2026adaponderlm,jaber2026tide}.

On the interpretability side, the limitations of using attention weights alone to explain token-to-token interaction have also been emphasized. Mohebbi et al.\ propose a quantification of context mixing at the block level \cite{mohebbi2023contextmixing}. Fernando and Guitchounts treat the residual stream as a dynamical system evolving across depth and discuss attractor-like behavior in low-dimensional projections and layer-wise trajectories \cite{fernando2025transformerdynamics}. These works are closely aligned with the present study in treating a Transformer as more than a mere stack of layers. However, they do not provide a framework that separates the attention and MLP terms and then operationally measures basin selection and subsequent stabilization through intervention-response analysis.

The stance of the present work is to interpret differences in token-wise computational depth not merely as a target for efficiency optimization, but as a consequence of a division of roles between basin selection by Self-Attention and in-basin stabilization or refinement by the MLP.
For this purpose, we introduce the interventions \emph{attention stopping} and \emph{MLP stopping}, and
we operationally characterize basin-like selection by examining whether output sensitivity to attention-side suppression is reduced at later intervention onset depths.
In this sense, the present work makes contact with both token-wise adaptive-depth research and dynamical-systems analyses of the residual stream, while distinguishing itself by connecting the two through the perspectives of intervention, output invariance, and component-wise role separation.

Taken together, these lines of work show that internal computation can be read out, intervened upon, and in some cases interpreted dynamically, but they differ substantially in what they measure, at what resolution, and whether they explicitly separate the roles of Self-Attention and the MLP. This distinction is summarized more directly in Table~\ref{tab:comparison_prior_methods_compact}.

\subsection{Dynamical Systems Formulation}

In this work, we regard the internal state of a Transformer as a point on a high-dimensional sequence-level state space $\mathcal{H}$ and describe the depth-wise computation as a discrete-time dynamical system:
\[
H^{(l+1)} = F^{(l)}\!\left(H^{(l)}\right).
\]

Within this framework, we introduce the following concepts:
\begin{itemize}

\item \textbf{Basin}: an operational output-associated region, namely a set of sequence states corresponding to the same output, or more weakly the same output distribution, under a specified output criterion;

\item \textbf{Invariance region}: a region in which the output remains effectively invariant under suppression of the attention contribution,
\item \textbf{Regime}: a class of dynamical behavior observed at a given depth range.
\end{itemize}

It is important to note that the present dynamical viewpoint is operational and finite-horizon. The update map depends explicitly on depth through the layer index \(l\), so the system is non-autonomous rather than governed by a single time-invariant flow. Accordingly, the basin-like and invariance-region notions introduced here are not intended as global attractor basins in the strict sense of autonomous dynamical systems. Instead, they should be understood as depth-indexed operational notions defined with respect to the remaining sequence of updates \(\{F^{(l)}\}_{l \ge k}\) in the trained network.

We do not claim that the basin-based picture is the unique correct description; rather, we introduce it as a useful organizing framework for the observed dynamics.

\subsection{Positioning of This Work}

Table~\ref{tab:comparison_prior_methods_compact} summarizes the relation between the present work and several closely related lines of research. The key point is not that internal Transformer computation has never been probed quantitatively, nor that depth-wise non-uniformity has never been observed. Rather, the distinguishing feature of the present work is to organize probing and intervention around token-level output invariance under a residual decomposition that explicitly separates Self-Attention and MLP contributions. A more detailed comparison is provided in Appendix Table~\ref{tab:comparison_prior_methods_full}.

\begin{table*}[t]
\centering
\footnotesize
\setlength{\tabcolsep}{4pt}
\renewcommand{\arraystretch}{1.10}
\begin{tabularx}{\textwidth}{
>{\raggedright\arraybackslash}p{2.55cm}
>{\raggedright\arraybackslash}p{3.05cm}
X}
\toprule
\textbf{Line of work} & \textbf{Representative papers} & \textbf{Relation to the present work} \\
\midrule \midrule

Readout-based probing
&
Logit Lens, Tuned Lens \cite{nostalgebraist2020logitlens,belrose2023tunedlens}
&
Shows that intermediate predictions can be read out quantitatively, but does not define intervention-based output invariance.
\\ \hline

Causal intervention / editing
&
ROME \cite{meng2022rome}
&
Demonstrates causal localization of internal computation, but not token-wise effective depth under sustained layer suppression.
\\ \hline

Late-layer redundancy / depth-wise non-uniformity
&
\cite{benartzy2024attendfirst,he2024notallattention,lad2024stages}
&
Provides the closest empirical precedents for late-layer attention redundancy and stage-wise depth sensitivity;
our work differs by defining token-level output invariance and characterizing token-wise stabilization through intervention-response curves.
\\ \hline

Dynamical-systems viewpoints
&
\cite{lu2019mpds,geshkovski2023clusters,geshkovski2023mathperspective}
&
Provides theoretical dynamical interpretations of Transformer computation, but not an empirical SA/MLP decomposition with token-wise output invariance under intervention.
\\ \hline

\textbf{This work}
&
---
&
Organizes probing and intervention around token-level output-side stability under
an explicit residual decomposition into Self-Attention and MLP contributions.
\\

\bottomrule
\end{tabularx}
\caption{Compact positioning of the present work relative to closely related lines of research. A more detailed comparison is provided in Appendix Table~\ref{tab:comparison_prior_methods_full}.}
\label{tab:comparison_prior_methods_compact}
\end{table*}

\section{Transformer as a Structured Discrete-Time Dynamical System}
\label{sec:struct_DDS}

In this section, we formulate depth-wise Transformer inference as a discrete-time dynamical system on the state space of the full token sequence, and then introduce a residual decomposition that separates the Self-Attention contribution from the MLP contribution. The purpose of this section is threefold. First, we make explicit the level of description adopted in this paper: the dynamical state is the representation of the entire sequence, not that of an individual token. Second, we present a component-wise decomposition that supports the intervention-based measurements used in the later sections. Third, we define the output-based criterion of effective invariance that underlies our notions of basin selection, invariance region, and token-wise critical depth.

\subsection{Sequence-level state and discrete-time formulation}

Let $n$ denote the number of tokens in the input sequence and $d$ the hidden dimension of each token representation. For each layer $l$ and token index $i \in \{1,\dots,n\}$, let
\[
 h_i^{(l)} \in \mathbb{R}^{d}
\]
denote the hidden representation of token $i$ at layer $l$. We define the \emph{sequence-level state} at layer $l$ by
\begin{equation}
H^{(l)} = \left(h_1^{(l)},\dots,h_n^{(l)}\right) \in \mathcal{H},
\label{eq:sequence_state_def}
\end{equation}
where the state space is identified as
\begin{equation}
\mathcal{H} \cong (\mathbb{R}^{d})^{n} \cong \mathbb{R}^{n \times d}.
\label{eq:state_space_identification}
\end{equation}
Thus, throughout this paper, the term \emph{state} refers to the representation of the entire token sequence. Individual token representations are treated as components of that state. This distinction is important because Self-Attention acts through couplings among tokens, whereas the MLP acts token-wise on each component.

With this notation, the forward computation of a Transformer across depth can be written as a discrete-time dynamical system
\begin{equation}
H^{(l+1)} = F^{(l)}\!\left(H^{(l)}\right),
\label{eq:full_block_map}
\end{equation}
where $F^{(l)} : \mathcal{H} \to \mathcal{H}$ denotes the full transformation implemented by layer $l$, including the residual pathway and internal submodules. For deep residual models, it is standard to rewrite the update in increment form and relate it, in a suitable continuous-depth limit, to an ordinary differential equation \cite{haber2017resnetODE,chen2018neuralode}. In the present discrete setting, we write
\begin{equation}
H^{(l+1)} = H^{(l)} + \Phi^{(l)}\!\left(H^{(l)}\right),
\label{eq:increment_form}
\end{equation}
where $\Phi^{(l)}$ represents the net residual increment generated at layer ${l}$. We use this form only as an organizational device for inference-time computation; we do not assume that a continuous-time limit is required for the subsequent analysis.

It is also important to emphasize that the update map depends explicitly on the layer index $l$. In this sense, the present formulation is \emph{non-autonomous}: the dynamics are governed by a depth-indexed family of maps $\{F^{(l)}\}$ rather than by a single time-invariant flow. This point matters when interpreting basin-like or invariance-region notions later in the paper.

\subsection{Residual decomposition into Self-Attention and MLP contributions}

The core methodological step of this work is to decompose the residual increment into two observable contributions, corresponding to Self-Attention and the MLP. At an abstract level, independent of a particular implementation detail, we write
\begin{equation}
H^{(l+1)}
=
H^{(l)}
+
\Delta_A^{(l)}\!\left(H^{(l)}\right)
+
\Delta_M^{(l)}\!\left(H^{(l)};\mathcal{C}^{(l)}\right),
\label{eq:general_dual_decomp}
\end{equation}
where $\Delta_A^{(l)}$ denotes the Self-Attention contribution and $\Delta_M^{(l)}$ the MLP contribution. The auxiliary quantity $\mathcal{C}^{(l)}$ is included to accommodate architecture-dependent internal wiring. In a sequential Pre-Norm block, it may represent the intermediate state produced after the attention sublayer; in a parallel residual formulation, this extra argument is unnecessary.

This notation is intended to separate the residual update into two conceptually distinct components before committing to a specific block implementation. It also clarifies the level at which intervention is defined: the interventions considered in this paper act on additive \emph{contributions to the sequence-level update}, not on isolated token coordinates in abstraction from the rest of the sequence.

Under this view, the interventions introduced later in the paper are expressed as selective suppression of one of the additive components beyond a specified depth. Schematically, if one component is suppressed, the update takes the form
\begin{equation}
H^{(l+1)}
=
H^{(l)}
+
\Delta_{\mathrm{keep}}^{(l)},
\qquad
\text{(one component suppressed)},
\label{eq:general_intervention_view}
\end{equation}
where $\Delta_{\mathrm{keep}}^{(l)}$ denotes the contribution that remains active. In this sense, the present framework is a general method for analyzing \emph{component-wise intervention and response}. It applies both to implementations in which Self-Attention and the MLP are connected sequentially within a layer and to implementations in which both read the same state and are added in parallel.

In the experiments of this paper, we study Pythia-410M \cite{biderman2023pythia}, which is based on a GPT-NeoX-style architecture. For this model, the layer update is naturally written as the sum of two additive contributions evaluated on the same sequence-level state:
\begin{equation}
H^{(l+1)}
=
H^{(l)}
+
A^{(l)}\!\left(H^{(l)}\right)
+
M^{(l)}\!\left(H^{(l)}\right),
\label{eq:pythia_parallel_dual}
\end{equation}
where $A^{(l)}$ is the Self-Attention update term and $M^{(l)}$ is the MLP update term. Thus, in the concrete setting of Pythia, both update components read the same layer state and contribute additively to the residual stream.

We emphasize that the main point of the present paper is not the architectural distinction between sequential and parallel implementations as such. Rather, it is to provide an analysis framework in which the attention contribution and the MLP contribution can be treated as separable, intervenable components of the sequence-level dynamics. For this reason, the subsequent empirical sections adopt the concrete formulation in Eq.~\eqref{eq:pythia_parallel_dual}, while the conceptual structure of the analysis is intended to remain applicable to more general Transformer variants.

\subsection{Output-space criterion for effective invariance}
\label{sec:output_invariance_criterion}

The intervention analysis in later sections is ultimately evaluated at the level of the model output. Let
\[
p_{\mathrm{full}}^{(t)}
\]
denote the clean output distribution associated with token position $t$ under ordinary inference, and let
\[
p_{\mathrm{int}}^{(t;k)}
\]
denote the corresponding output distribution under a specified intervention that is activated from depth $k$ onward. Here $t$ indexes the token whose output behavior is being evaluated, and $k$ denotes the intervention onset depth.

To quantify the effect of intervention on the output, we introduce an output-difference functional
\begin{equation}
D_{\mathrm{out}}\!\left(p_{\mathrm{full}}^{(t)},\, p_{\mathrm{int}}^{(t;k)}\right),
\label{eq:output_distance_general}
\end{equation}
where the exact choice of distance measure is experiment-dependent. In the clearest case, one may take
\begin{equation}
D_{\mathrm{out}}\!\left(p,q\right)=D_{\mathrm{KL}}(p\|q),
\label{eq:output_distance_kl}
\end{equation}
but other output-space distances such as $\ell_1$ may also be used when explicitly stated. The key point is that the intervention is judged not by hidden-state proximity, but by whether the induced output remains effectively unchanged.

We therefore say that the output for token $t$ is \emph{effectively invariant} under an intervention activated from depth $k$ if
\begin{equation}
D_{\mathrm{out}}\!\left(p_{\mathrm{full}}^{(t)},\, p_{\mathrm{int}}^{(t;k)}\right)\le \varepsilon,
\label{eq:effective_invariance_def}
\end{equation}
where $\varepsilon > 0$ is a prescribed tolerance. This tolerance is part of the operational definition and should be understood as a measurement choice rather than an architecture-independent constant.

Based on Eq.~\eqref{eq:effective_invariance_def}, we define the \emph{token-wise critical depth} by
\begin{align}
k_{\mathrm{crit}}(t) &\equiv k^\ast(t) \nonumber \\
&= 
\min \Bigl\{
k \in \{0,\dots,L-1\}
\;\Big|\;
D_{\mathrm{out}}\!\left(p_{\mathrm{full}}^{(t)},\, p_{\mathrm{int}}^{(t;k')}\right)\le \varepsilon
\ \text{for all } k' \ge k
\Bigr\},
\label{eq:tokenwise_critical_depth}
\end{align}
whenever such a depth exists. Intuitively, $k_{\mathrm{crit}}(t)$ is the earliest depth beyond which activating the specified intervention no longer changes the output of token $t$ beyond the tolerance $\varepsilon$.

This definition is intentionally operational. It does not assert that the hidden state has converged in a metric on $\mathcal{H}$, nor that the trajectory has reached a fixed point in the strict dynamical-systems sense. Rather, it identifies a depth beyond which the remaining depth-wise computation becomes effectively irrelevant to the output under the intervention criterion being used.

The empirical instantiation of this criterion may use either a
distribution-level distance or lower-dimensional output-side diagnostics,
depending on what is recorded in a given intervention experiment.

\subsection{Operational notions of basin and invariance region}

The output-based criterion above motivates the operational notions used later in the paper.
Informally, what we call a \emph{basin} is an operational, output-associated region of the sequence-level state space: a set of sequence states that remain compatible with the same output, or more weakly with the same output distribution, under the chosen output criterion.
An \emph{invariance region} is a subset of state space in which the output remains effectively unchanged under the corresponding intervention. Neither notion is intended as a canonical geometric object independent of measurement choice; both depend on the output distance functional and the tolerance used to judge effective invariance.

Accordingly, the basin-like and invariance-region notions employed in this paper should be read as \emph{depth-indexed operational notions} with respect to the remaining update sequence $\{F^{(l)}\}_{l\ge k}$ in the trained network. Their role is to provide a compact language for describing when sequence-level reorganization appears to have effectively completed, and when the remaining computation is better interpreted as local refinement within an already selected output-associated region.

\section{Dynamical Roles of Self-Attention and MLP}\label{sec:DRofSAandMLP}
Based on the sequence-level formulation introduced in Section~\ref{sec:struct_DDS},
this section analyzes the geometric and dynamical roles of Self-Attention and the MLP.
Our central claim is that these two update components act differently on the sequence-level state and that this heterogeneity gives the depth-wise dynamics their structure.

\subsection{Self-Attention as Global Reorganization of Token Representations}
Let $X \in \mathbb{R}^{n\times d}$ denote a sequence state written in matrix form, with row $x_i \in \mathbb{R}^d$ representing the hidden state of token $i$.
Ignoring the multi-head decomposition and output projection for the moment, a Self-Attention update can be written schematically as
\begin{equation}
\mathrm{Attn}(X)=P(X)X,
\end{equation}
where $P(X) \in \mathbb{R}^{n\times n}$ is a row-normalized attention matrix whose $(i,j)$ entry gives the weight from token $j$ to token $i$.
This expression is schematic and suppresses the value and output projections;
it is used only to emphasize the sequence-level mixing structure of Self-Attention.
More precisely, at each query position, the attention update is formed from a weighted
combination of projected value vectors associated with the current sequence, followed
by an output projection. Its defining feature is therefore cross-token mixing rather than
independent position-wise transformation.

This observation motivates the following interpretation: Self-Attention does not
transform each token representation independently; rather, it updates each position
through interactions with representations at other accessible positions in the sequence.
In dynamical terms, its primary role is to reshape inter-token relations and, through
that reshaping, to steer the sequence state toward a particular region of the state space.
For this reason, we interpret the Self-Attention contribution as a mechanism of sequence-level
reorganization and direction selection.

\subsection{MLP as Token-Wise Local Drift}
The MLP submodule, by contrast, acts independently on each token.
For a single token representation $x \in \mathbb{R}^d$, we write
\begin{equation}
M(x)=W_2\,\sigma(W_1x+b_1)+b_2,
\end{equation}
where $W_1 \in \mathbb{R}^{d_{\mathrm{ff}} \times d}$ and $W_2 \in \mathbb{R}^{d \times d_{\mathrm{ff}}}$ are learned weight matrices, $b_1 \in \mathbb{R}^{d_{\mathrm{ff}}}$ and $b_2 \in \mathbb{R}^{d}$ are bias vectors, $d_{\mathrm{ff}}$ denotes the intermediate MLP dimension, and $\sigma$ denotes the point-wise nonlinearity.
This map depends only on the state of the token to which it is applied; it does not directly mix information across distinct token indices.

Accordingly, the MLP does not directly exchange information across token positions,
although its position-wise nonlinear transformations can still alter the collective
geometry of the token representations. Instead, it transforms each token representation
independently, operating on states whose evolution has already been influenced by
attention-mediated cross-token interactions.
This motivates interpreting the MLP contribution as a token-wise local drift field.
More specifically, it refines individual token states within a sequence-level trajectory
that is jointly shaped across depth by attention-mediated cross-token interactions.

\subsection{Emergent Structure of the Layer-Wise Dynamics}
Combining the two update components, the sequence-level Transformer dynamics in the experimental setting of this paper are written as
\begin{equation}
H^{(l+1)}=H^{(l)}+A^{(l)}\!\left(H^{(l)}\right)+M^{(l)}\!\left(H^{(l)}\right).
\end{equation}
This update is the superposition of two qualitatively different actions:
\begin{itemize}
\item Self-Attention: a global operation that reorganizes the arrangement and coupling structure of the full token sequence,
\item MLP: a token-wise local operation that deforms each component of the sequence state individually.
\end{itemize}
Because these two actions differ in scale and in mode of interaction, the depth-wise computation is not naturally interpreted as a single homogeneous transformation.
Rather, each layer superimposes two distinct update mechanisms, and the resulting dynamics accumulate across depth.
From this viewpoint, Self-Attention primarily contributes to sequence-level direction selection, whereas the MLP primarily supports local refinement and stabilization once that direction has been selected.

\subsection{Basin--Drift Interpretation}
The preceding discussion suggests a dynamical interpretation of the depth-wise sequence-state trajectory.
Recall that the state at layer $l$ is
\begin{equation}
H^{(l)}=\left(h_1^{(l)},\dots,h_n^{(l)}\right)\in\mathcal{H},
\end{equation}
with $h_i^{(l)}\in\mathbb{R}^d$ and $\mathcal{H}\cong(\mathbb{R}^d)^n\cong\mathbb{R}^{n\times d}$.
Thus, the dynamical variable is the full token sequence, not an isolated token.

Since Self-Attention acts on the whole sequence state, it naturally contributes a global update that can move the trajectory across regions of $\mathcal{H}$.
By contrast, because the MLP applies token-wise nonlinear maps, it contributes local deformations on the components of the state once a sequence-level direction has been established.
This heterogeneous update structure is consistent with a picture in which the state space contains multiple output-associated regions, and the trajectory is first guided toward one such region and then progressively stabilized within it.

In this paper, we use the term \emph{basin} in this operational sense: it denotes a compact label for an output-associated region of the sequence-level state space, defined with respect to the chosen output criterion and observational tolerance.
This is not intended as a rigorous basin-of-attraction concept in the classical dynamical-systems sense.
Rather, it is a measurable notion introduced to organize intervention-response observations during inference.
Under this interpretation, basin selection is a sequence-level process, because the relevant state variable is the full token configuration.

At the same time, the token-wise components of the state need not stabilize uniformly.
Different tokens may evolve at different rates, and the refinement inside an output-associated basin may proceed asynchronously across components.
The resulting picture is therefore a hierarchical dynamical system in which sequence-level region selection and token-wise local stabilization proceed together but not necessarily on the same time scale.
We emphasize again that this is offered as an effective interpretive model of the observed dynamics, not as a claim that a unique or strict convergence structure has been mathematically established.

\subsection{Implications for Observable Dynamics}
If the above interpretation is appropriate, then the following observable tendencies should appear in depth-wise Transformer inference:
\begin{itemize}

\item the sensitivity of measured output-side diagnostics to further
context-side attention-output suppression should diminish after
basin-like selection is effectively completed;

\item the MLP contribution should become relatively more important in the later layers,
\item different tokens should exhibit different stabilization time scales and different depth-wise trajectories.
\end{itemize}
These tendencies can be examined experimentally through intervention-response measurements, changes in the output distribution, and descriptive indicators such as attention entropy.
In the next section, we test these predictions operationally on a trained GPT-style Transformer.

\section{Empirical Validation of Structured Dynamics}
\label{sec:experiment_early_exit}

In this section, we examine whether the dynamical predictions derived above are
consistent with the behavior of an actual trained Transformer during inference.
Our aim is not to improve benchmark performance, but to construct an operational
measurement framework for sequence-level depth-wise dynamics.

\subsection{Experimental Setup}
\label{sec:exp_setup}

To test the hypotheses developed in Section~4---namely, regime transitions
implied by the decomposition into Self-Attention and MLP contributions,
finite-depth reduction of intervention sensitivity associated with basin-like selection,
and token-dependent
non-uniform stabilization---we conduct intervention-based inference-time
experiments on a trained GPT-style Transformer. The guiding idea is to treat
selective suppression of an additive update component as a measurement operation
on the sequence-level dynamics.

\paragraph{Model.}
As the target model we use a decoder-only autoregressive Transformer
(GPT-style). Specifically, we adopt \texttt{Pythia-410M}
\cite{biderman2023pythia} from the Pythia series released by EleutherAI.
This model has $L=24$ layers, hidden dimension $d=1024$, and $H=16$ attention
heads. Because the structural hyperparameters are fixed and publicly documented,
the resulting analysis is highly reproducible. Inference is performed
deterministically: training-time stochastic elements such as dropout are
disabled, and identical inputs yield identical internal state sequences.

\paragraph{Prompt / token roles.}
The input sequence is analyzed by distinguishing (i) context-side token
positions and (ii) the output/readout token position. The context-side tokens
are treated as components of the sequence state that participate in
sequence-level reorganization and basin-like selection, whereas the output token
is treated as the readout location associated with the final next-token
prediction. In the figures below, for representative prompts, we separately
analyze specific context-side tokens (e.g., \texttt{recognition},
\texttt{symbolic}) and the output token. 
We have also confirmed qualitative reproducibility across multiple prompts, although exhaustive visualization
across prompts is left to future work.

For a selected token position $t$, the final-layer representation $h_t^{(L)}$ defines
a next-token distribution conditioned on the prefix ending at that position.
This readout is not a reconstruction of the input token $x_t$. Although ordinary
autoregressive generation uses only the distribution at the final prompt position
to select the next token, the intermediate-position distributions remain valid
prefix-conditioned predictions. We use their changes under intervention as
diagnostic measures of token-position-dependent stability.

\paragraph{Representative prompts.}
To illustrate input dependence, we present results for the following two
prompts:
\begin{itemize}
  \item \textbf{Prompt 1}: \texttt{It would be a good idea to use this technique when}
  \item \textbf{Prompt 2}: \texttt{While neural networks excel at pattern recognition, symbolic methods}
\end{itemize}
In the figures and main text, we refer to these simply as \textbf{Prompt 1}
and \textbf{Prompt 2}.

\paragraph{Intervention: attention-side suppression.} 
To examine the dynamical role of self-attention across depth, we suppress the additive attention-output
contribution beyond a selected layer $k$. For historical continuity with the implementation, we refer to
this operation as attention stopping, although it is operationally an attention-side suppression applied
at the attention-output hook.

In a decoder-only GPT-style model, two positional settings can be considered for attention-side suppression. All-token stopping (A-stop) suppresses the attention-output contribution at all token positions, including the final position, whereas context-side stopping (C-stop) excludes the final position used for next-token readout. Thus, A-stop and C-stop differ in the positional scope of the attention-side intervention.

The distinction may itself be informative, because suppression at the final readout position can affect the output more directly than suppression at the preceding context positions. However, the magnitude and depth profile of this difference are likely to depend strongly on the prompt and the dependencies among its tokens. A systematic comparison between A-stop and C-stop is therefore left for future work. The context-side analyses in Sections~5.2 and~5.3 report only the C-stop results.

Specifically, for layers $\ell \geq k$, C-stop suppresses the attention-output contribution at all context positions while retaining it at the final readout position. Thus, further attention-mediated updates at the context positions are suppressed, whereas attention-mediated processing at the final readout position, the residual pathway, and the MLP contribution remain active. This intervention allows us to examine changes in the sequence-level trajectory without directly suppressing computation at the position from which the next-token distribution is read out.

In our implementation, hook-based interventions on internal activations and cache retrieval during
inference are realized using TransformerLens~\cite{nanda2022transformerlens_ext}.

\paragraph{Intervention: MLP suppression.}
In parallel with attention-side suppression, we also introduce MLP-stop, which
suppresses the additive MLP contribution beyond layer $k$. Concretely, for
layers $l \ge k$, we replace the sequence-level update
\[
H^{(l+1)}
=
H^{(l)}
+
A^{(l)}\!\left(H^{(l)}\right)
+
M^{(l)}\!\left(H^{(l)}\right)
\]
by an update in which the MLP contribution is removed while the attention
contribution and the residual pathway are preserved:
\[
H^{(l+1)}
=
H^{(l)}
+
A^{(l)}\!\left(H^{(l)}\right).
\]
As a measurement operation paired with attention-side suppression, this allows
us to examine whether the MLP behaves as a token-wise local drift component
along depth. In Section~\ref{sec:output_readout}, we compare MLP-stop and
ATTN-stop for the output token.

\paragraph{Layer sweep and baseline.}
For each prompt, we first run the ordinary forward pass without intervention
and record the baseline output. We then sweep the intervention onset layer over
\[
k=0,1,\ldots,L-1,
\]
where $L=24$ for Pythia-410M, and apply the corresponding component suppression
from layer $k$ through the final layer. The ordinary forward pass without
intervention is reported separately as the baseline.

\paragraph{Measured quantities and aggregation.}
For the context-side analyses in Figs.~1 and~2, we apply the model's final
vocabulary readout to the final-layer representation at each selected token position.
The resulting logits define a next-token distribution conditioned on the prefix ending
at that position. Our primary interest is not the predicted token itself, but the change
in this prefix-conditioned distribution when the attention-side intervention is activated from different depths.

The output-space criterion introduced in Section~3.3 provides a general
operational definition of effective invariance in terms of an output-distance
functional $D_{\mathrm{out}}$ and a tolerance $\varepsilon$. In the exploratory
experiments reported here, however, we do not impose a fixed numerical tolerance
and we do not compute a full distribution-level KL divergence for the
intervention sweep. Instead, we use output-side scalar diagnostics computed from
the output distribution, and we inspect their intervention-response curves
directly.

For the context-side analyses in Figs.~1 and~2, we report changes in position-wise
scalar diagnostics derived from the prefix-conditioned next-token distribution, relative
to the ordinary forward pass. The reduction of these intervention-response differences at
later onset layers is used as an empirical indication that the corresponding token-position
prediction has become less sensitive to the remaining attention-mediated updates.
This reduction provides an empirical guide to effective invariance and to token-wise
stabilization behavior across depth.
The calibration of a robust numerical tolerance or a model-independent
distribution-level distance criterion is left to future work involving broader experiments
across model sizes, architectures, prompts, and intervention variants.

To quantify the depth-wise dynamics, we measure the following quantities:
\begin{itemize}
  \item \textbf{Logit margin.}
    At a selected token position, we compute the difference between the top-1 and top-2
    logits of the prefix-conditioned next-token distribution, denoted by $\Delta_{\mathrm{logit}}$.
    This quantity measures the decisiveness of the position-wise diagnostic prediction.

\item \textbf{Prediction entropy.}
At a selected token position, we compute the entropy of the prefix-conditioned
next-token distribution $p(\cdot)$: \[ H_{\mathrm{pred}} = -\sum_i p_i \log p_i, \] 
where $p_i$ is the predicted probability of vocabulary item $i$. 
This quantity measures the uncertainty of the position-wise diagnostic prediction.

  \item \textbf{Scalar prediction differences under suppression.}
  For the intervention-response curves in Figs.~\ref{fig:token_recognition}
  and \ref{fig:token_symbolic}, we use absolute differences in scalar prediction
  diagnostics relative to the ordinary forward pass. For example, if
  $m_{\mathrm{full}}^{(t)}$ and $H_{\mathrm{full}}^{(t)}$ denote the baseline
  logit margin and prediction entropy for token $t$, and
  $m_{\mathrm{C\text{-}stop}}^{(t;k)}$ and
  $H_{\mathrm{C\text{-}stop}}^{(t;k)}$ denote the corresponding values under
  C-stop from layer $k$, we examine
  \[
  \Delta_{\mathrm{margin}}^{(t)}(k)
  =
  \left|
  m_{\mathrm{full}}^{(t)}
  -
  m_{\mathrm{C\text{-}stop}}^{(t;k)}
  \right|,
  \]
  and
  \[
  \Delta_{\mathrm{entropy}}^{(t)}(k)
  =
  \left|
  H_{\mathrm{full}}^{(t)}
  -
  H_{\mathrm{C\text{-}stop}}^{(t;k)}
  \right|.
  \]
These quantities are descriptive position-wise proxies for intervention sensitivity
rather than full distribution-level divergences.

  \item \textbf{Rank of the baseline top-$1$ token.}
  In the readout-side comparison of Section~\ref{sec:output_readout}, we
  record the rank of the baseline top-$1$ token within the patched output
  distribution. This measures how well the original top candidate is preserved
  under intervention.

  \item \textbf{Attention entropy (token-wise).}
  For layer $l$ and head index $h \in \{1,\dots,H\}$, let $p^{(l,h)}$ denote
  the attention distribution emitted from the target token. We define
  \[
  H_{\text{attn}}^{(l,h)}
  =
  -\sum_j p_j^{(l,h)} \log p_j^{(l,h)},
  \]
  where the sum runs over token indices $j$. Lower values indicate stronger
  concentration of attention.
\end{itemize}

Here, $j$ indexes token positions in the sequence and $H$ denotes the number
of attention heads. Unless otherwise stated, we average
$H_{\text{attn}}^{(l,h)}$ across heads and report
\[
\mu^{(l)}
=
\frac{1}{H}\sum_h H_{\text{attn}}^{(l,h)}
\]
together with its standard deviation $\sigma^{(l)}$. This aggregation treats
cross-head variation not merely as noise, but as a possible signature of role
differentiation.

\paragraph{Qualitative assessment of effective invariance.}
In the general notation of Section~3.3, effective invariance can be defined by
a tolerance condition of the form
\[
D_{\mathrm{out}}\!\left(
p_{\mathrm{full}}^{(t)},
p_{\mathrm{int}}^{(t;k)}
\right)
\le \varepsilon.
\]
In the present exploratory experiments, however, we do not use a fixed numerical
tolerance to determine $k_{\mathrm{crit}}(t)$. Instead, we visualize the full
intervention-response curves and use the reduction and relative stabilization
of the plotted differences at later onset layers as a qualitative indication of
token-wise stabilization. We therefore do not report numerically determined
values of $k_{\mathrm{crit}}(t)$ in the present study. The calibration of a
robust numerical tolerance is left to future work.

\paragraph{Theoretical correspondence.}
The experiments below examine the theoretical picture of Section~\ref{sec:DRofSAandMLP}
from three viewpoints: (i) finite-depth reduction in intervention sensitivity
associated with basin-like selection; (ii) regime-like organization, assessed
through changes in descriptive attention statistics such as $H_{\mathrm{attn}}$;
and (iii) token-wise asymmetry, assessed by comparing context-side and output-side behavior.
In this way, we seek an operational and quantitative description of
token-dependent and input-dependent sequence-level dynamics that are difficult
to recover from a purely averaged layer-wise view.

\subsection{Token-Wise Stabilization and Basin-Like Selection}
\label{sec:basin_completion}

In this subsection, we examine whether the prefix-conditioned prediction at
an individual token position becomes operationally stable at finite depth. 
The key idea is that, if suppressing context-side attention-output contributions
beyond a layer $k$ no longer materially changes the position-wise diagnostic readout,
then the corresponding token component has become effectively insensitive to the
remaining attention-mediated updates under the present observation criterion.
We then use these token-wise observations to consider whether the sequence-level
trajectory admits an operational basin-like interpretation.

\paragraph{Operational definition of basin selection.}
We distinguish two stages of output stabilization:
\begin{itemize}
  \item \textbf{(i) Basin-like selection:}
  the stage at which the trajectory enters a region of the sequence-level state
  space associated with the same output, or more generally with effectively the
  same output distribution under an observational criterion.

  \item \textbf{(ii) In-basin stabilization:}
  the stage at which the state continues to deform and stabilize while
  remaining within that output-associated region.
\end{itemize}

In the abstract operational formulation of Section~3.3, basin-like selection
may be defined by a tolerance condition on an output-distance functional.
In the present experiments, we use this condition as an operational idealization
and inspect the observed intervention-response curves directly.
A sustained reduction in the plotted intervention-response differences at
later onset layers is used as qualitative evidence of reduced sensitivity
to the remaining attention-mediated updates.

\begin{figure}[htb]
\centering
\begin{subfigure}{0.45\linewidth}
\includegraphics[width=\linewidth]{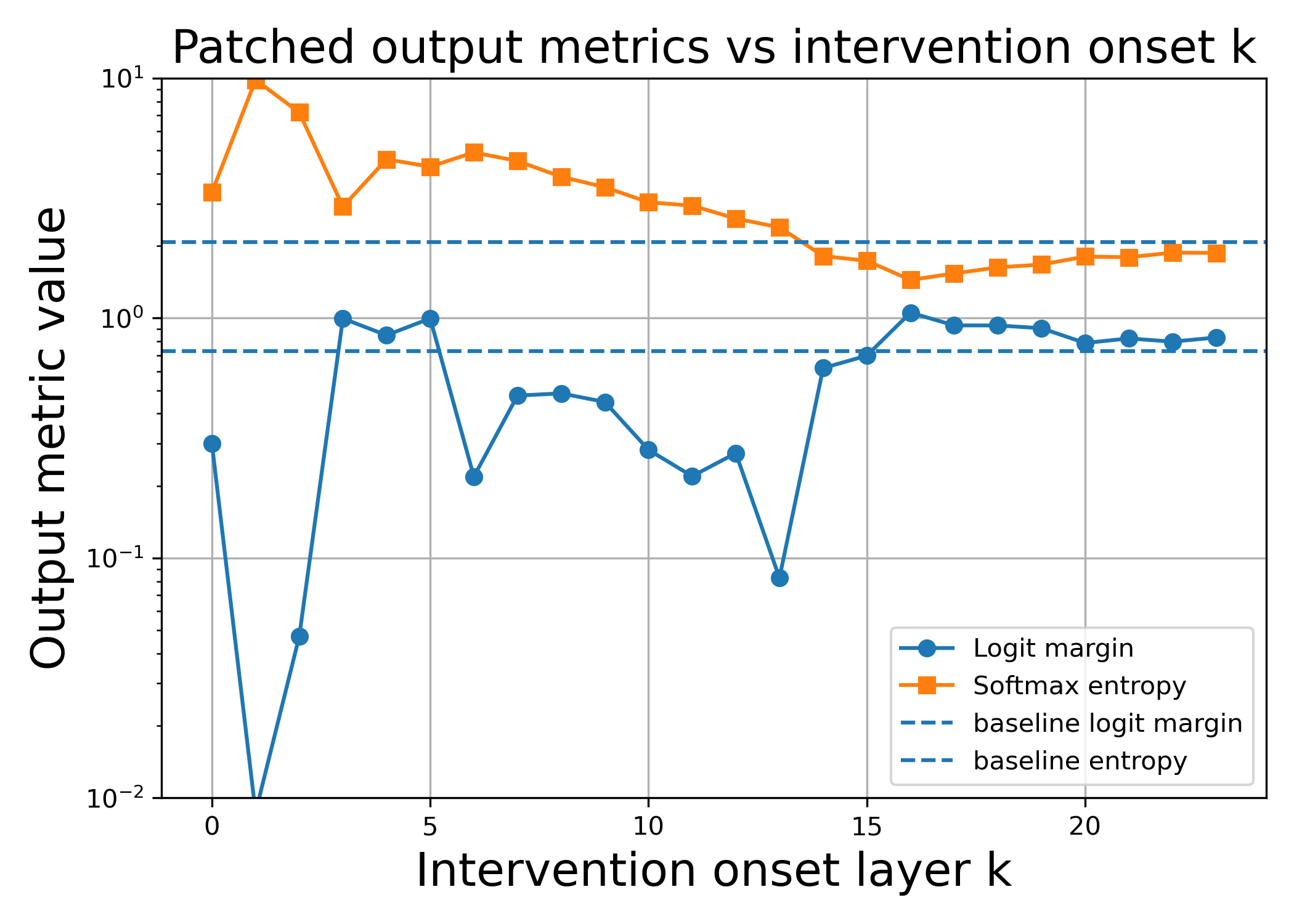}
\caption{}
\end{subfigure}
\begin{subfigure}{0.45\linewidth}
\includegraphics[width=\linewidth]{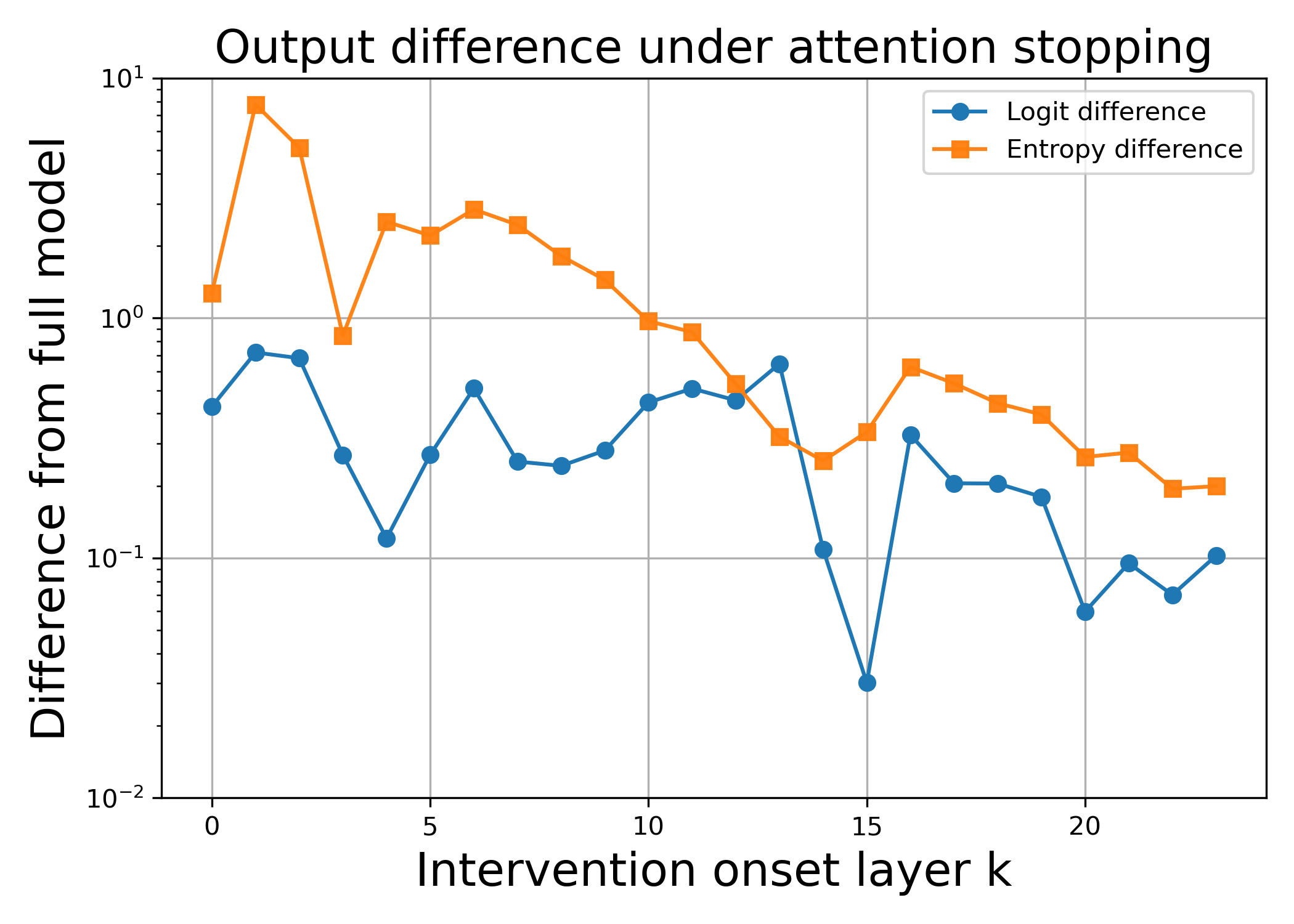}
\caption{}
\end{subfigure}

\caption{
Intervention-response behavior for the prefix ending at the token ``recognition.''
(a) Logit margin and softmax entropy of the next-token distribution read from the
final-layer representation at the selected token position, plotted against the
intervention onset layer $k$. This distribution is conditioned on the prefix ending
at ``recognition'' and is not a reconstruction of the token itself.
Each point is obtained by suppressing the attention-output contribution at the
context positions from layer $k$ onward.
Dashed lines indicate the corresponding values obtained without intervention.
(b) Absolute deviations from the unmodified-model values, $|\Delta m(k)|$ and $|\Delta H(k)|$.
The generally smaller deviations for later intervention onsets indicate reduced
sensitivity of the prefix-conditioned prediction to the remaining attention-mediated
updates, although the approach to the baseline is not strictly monotonic. 
}
\label{fig:token_recognition}

\end{figure}

\begin{figure}[htb]
\centering
\begin{subfigure}{0.45\linewidth}
\includegraphics[width=\linewidth]{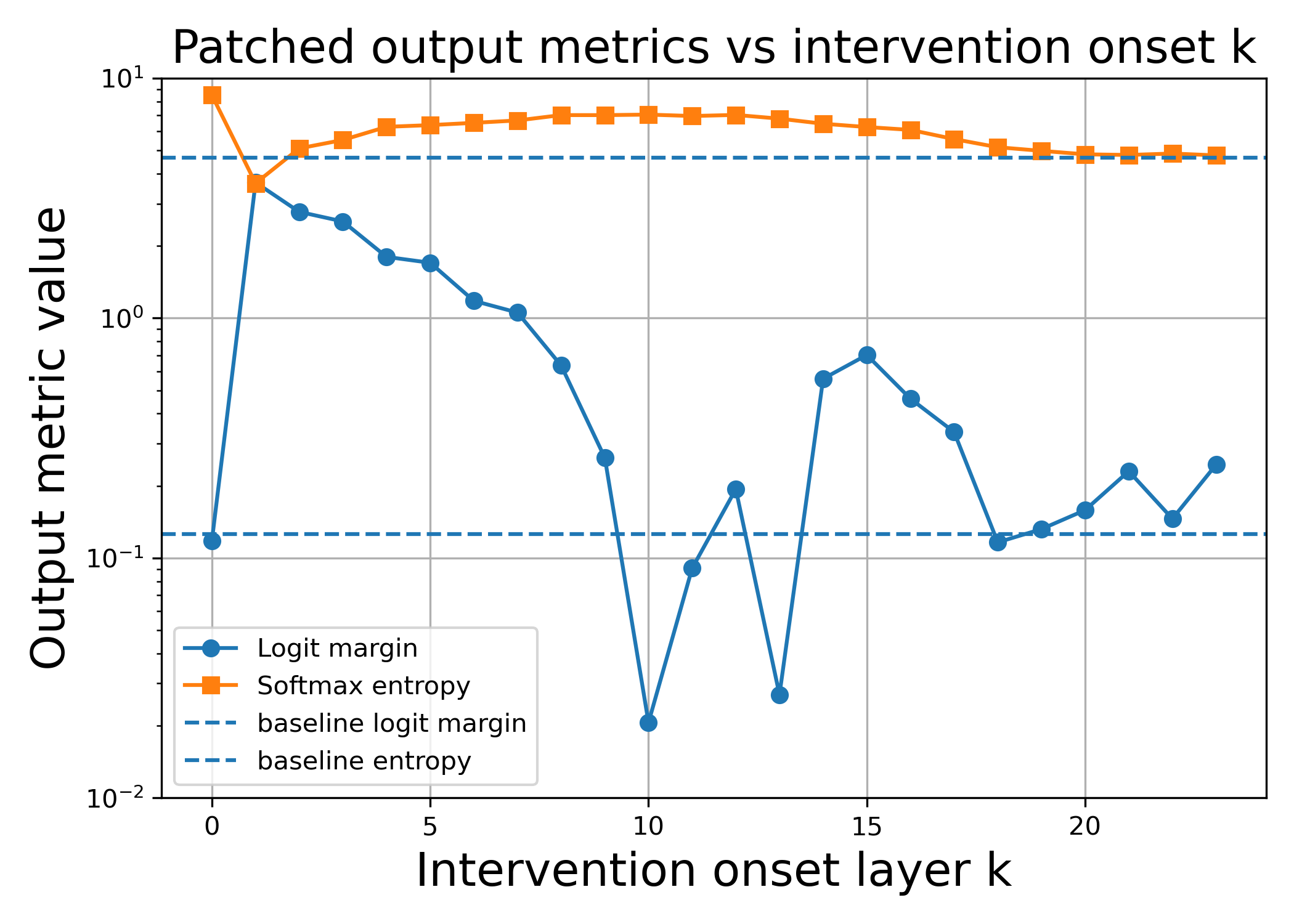}
\caption{}
\end{subfigure}
\begin{subfigure}{0.45\linewidth}
\includegraphics[width=\linewidth]{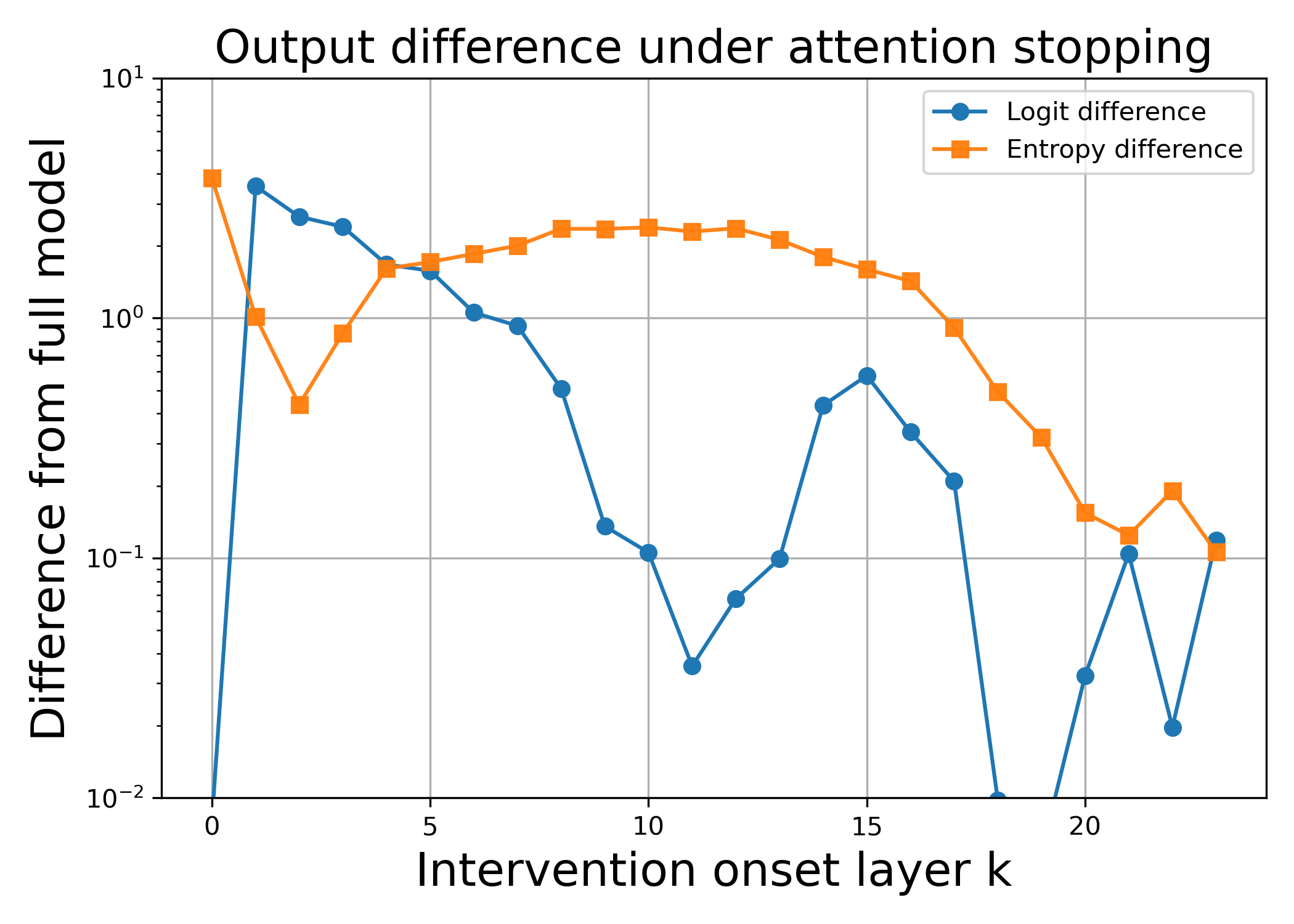}
\caption{}
\end{subfigure}

\caption{
Intervention-response behavior for the prefix ending at the token ``symbolic''
in the same prompt as Fig.~1. (a) Logit margin and softmax entropy of the
next-token distribution read from the final-layer representation at the selected
token position, plotted against the intervention onset layer $k$. The quantities
and intervention procedure are defined as in Fig.~1; dashed lines indicate the
corresponding values obtained without intervention.
(b) Absolute deviations from the unmodified-model values, $|\Delta m(k)|$
and $|\Delta H(k)|$. The deviations are generally reduced for later intervention
onsets, but the depth-wise response differs from that of the prefix
ending at ``recognition,'' illustrating token-position-dependent stabilization
of prefix-conditioned predictions.
}
\label{fig:token_symbolic}
\end{figure}

\paragraph{Observed stabilization of position-wise predictions.}
Figures~\ref{fig:token_recognition} and \ref{fig:token_symbolic} show the
results for two representative context-side tokens.
The numerical values underlying the intervention-response curves in Figs.~\ref{fig:token_recognition}--
\ref{fig:token_symbolic}  are listed in Appendix~\ref{sec:appendix_numerical_values}, 
Tables~\ref{tab:appendix_recognition_values} and~\ref{tab:appendix_symbolic_values}.
In both cases, the position-wise prediction diagnostics exhibit smaller deviations
from their unmodified-model values for later intervention onsets, indicating reduced
sensitivity to the remaining attention-mediated updates.

What matters most here is the intervention-response behavior. As the onset layer
$k$ increases, the scalar output differences under C-stop
tend to decrease and remain relatively small over a range of later intervention onset layers.
This indicates that the
sequence state reaches an output-associated region in which later context-side
attention-output suppression no longer materially alters the measured
position-wise prediction diagnostics.

\paragraph{Evidence consistent with finite-depth basin-like selection.}
Based on the operational interpretation above, the observed reduction of
the intervention-response differences at later onset layers suggests that
the selected token-position prediction becomes progressively less sensitive
to the remaining attention-mediated updates. This behavior is consistent with
basin-like selection occurring over a finite depth range, although the present
experiments do not determine a unique numerical completion depth.

This observation supports the following two-stage interpretation of the depth-wise dynamics:
\begin{itemize}
\item an earlier stage of competition among multiple candidate regions;
\item a later stage of reduced intervention sensitivity and stabilization within
an output-associated region.
\end{itemize}
Thus, the trajectory is not well described as a single undifferentiated
asymptotic stabilization process.

\paragraph{Token-dependent stabilization behavior.}
Comparing Figs.~\ref{fig:token_recognition} and
\ref{fig:token_symbolic}, 
we find that even under the same prompt, the apparent stabilization range and
the depth-wise intervention-response trajectory differ across token positions.
Output-side stabilization is therefore not synchronized across the sequence.
This supports the view that Transformer inference is a non-uniform dynamical
process, with distinct trajectories and stabilization time scales for different
components of the sequence-level state.

\paragraph{Interpretation in the dynamical-systems framework.}
These observations are consistent with the dynamical picture developed in
Section~4. Self-Attention contributes to global reorganization of the
sequence-level state and is therefore most relevant during basin-like selection.
Once the trajectory has entered an output-associated region, later layers can
continue to refine the state without materially altering the measured output
diagnostics, which is consistent with a relatively stronger role for the MLP
contribution in in-basin stabilization.

Together with the readout-side ATTN-stop/MLP-stop comparison in Section~5.4,
these observations are consistent with a transition from an earlier phase in
which attention-mediated sequence reorganization is more consequential to a later
phase in which MLP-mediated refinement plays a comparatively stronger role.

\subsection{Emergence of a Transition Structure in Attention Dynamics}
\label{sec:entropy_transition}

\begin{figure}[htb]
\centering
\begin{subfigure}{0.45\linewidth}
\includegraphics[width=\linewidth]{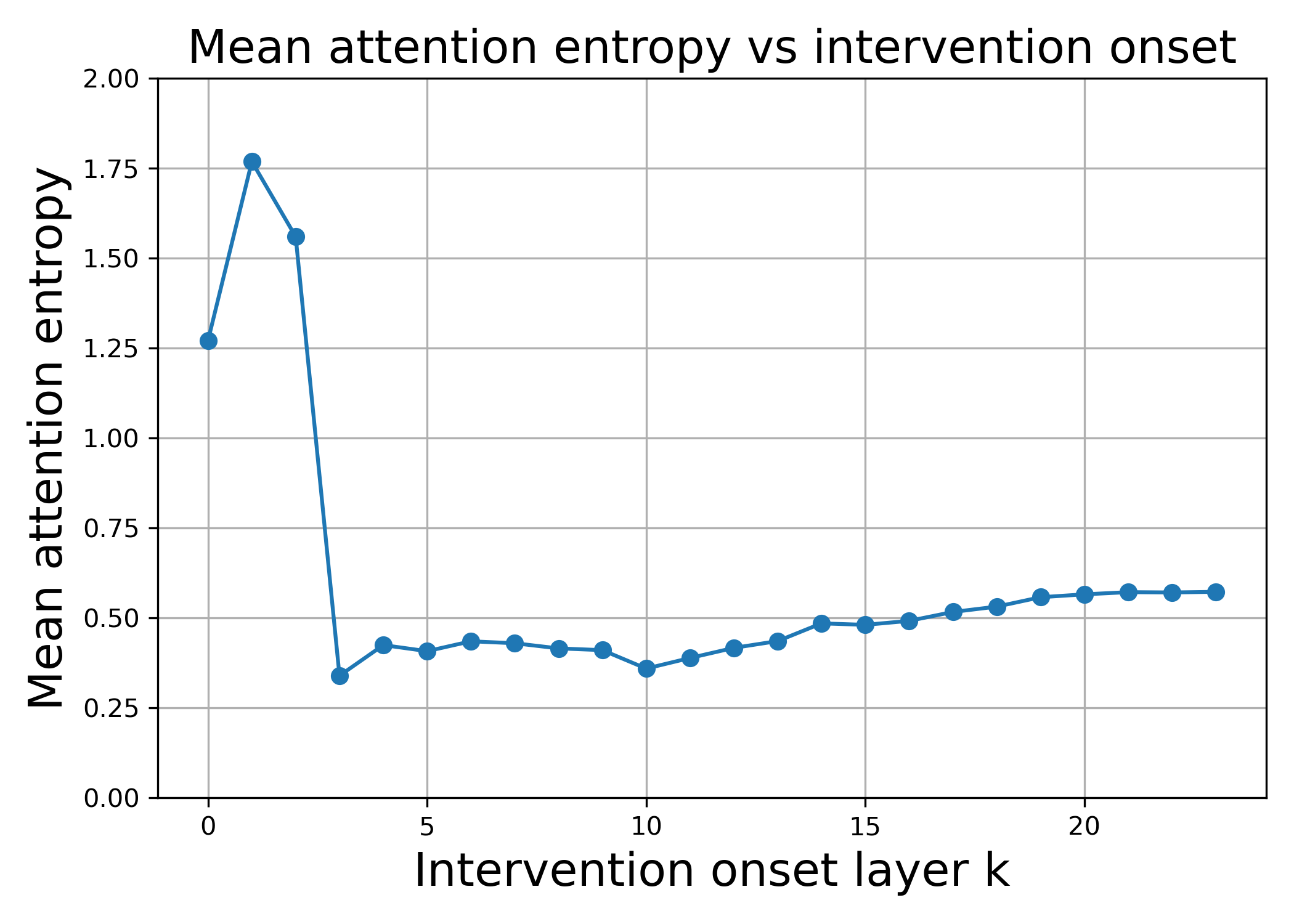}
\caption{}
\end{subfigure}
\begin{subfigure}{0.45\linewidth}
\includegraphics[width=\linewidth]{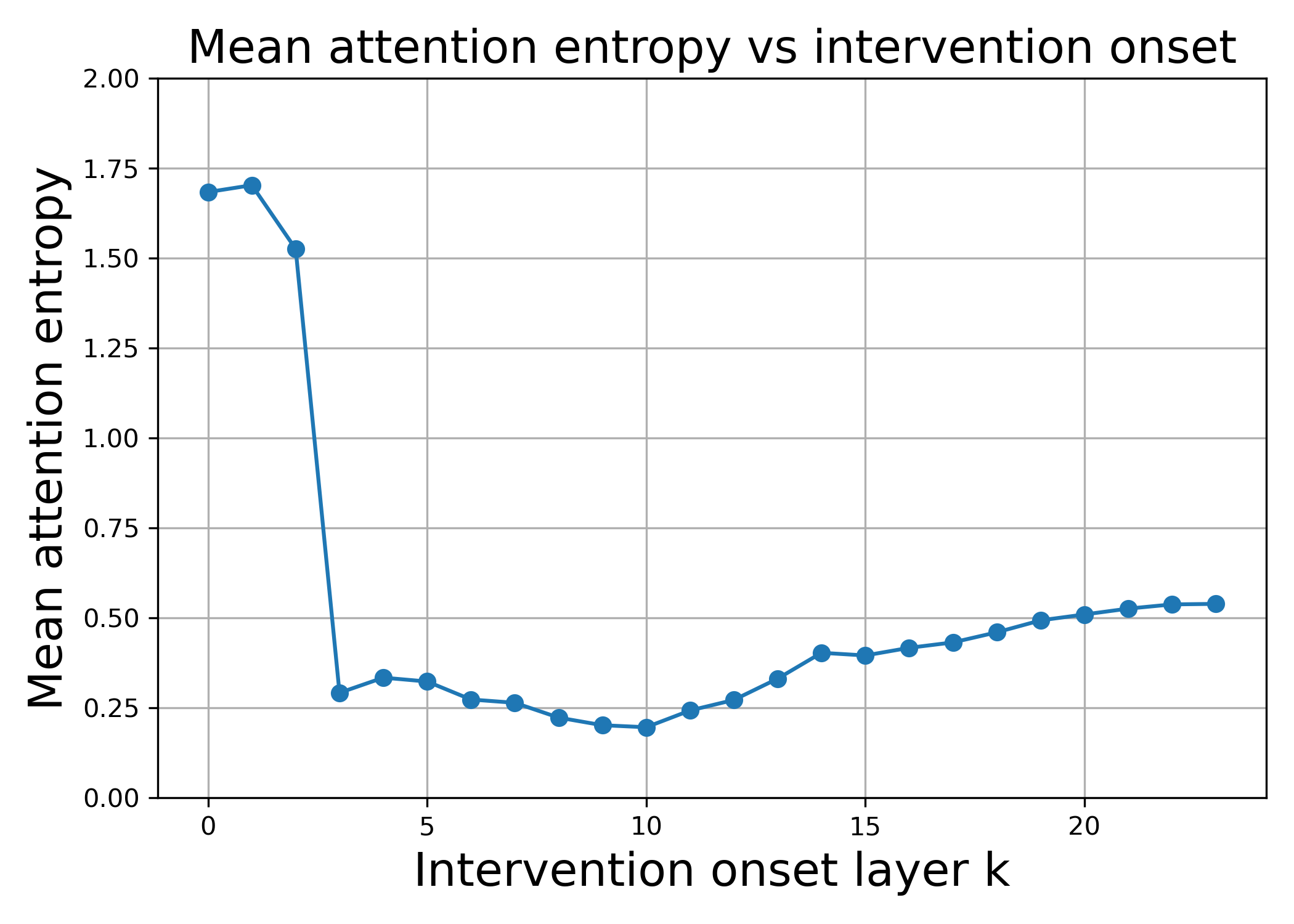}
\caption{}
\end{subfigure}
\caption{
Mean attention entropy for the tokens ``recognition'' and ``symbolic''
as a function of the intervention onset layer $k$. Panel~(a) shows ``recognition,''
and panel~(b) shows ``symbolic.'' For each $k$, entropy is first averaged over
attention heads at each layer and then averaged over the fixed layer range $\ell=12,\ldots,23$.
For ``recognition,'' the sharp decrease between $k=2$ and $k=3$ therefore
reflects a change in the recorded late-layer attention distributions rather
than a change in the set of layers included in the average.
For $k>12$, the fixed averaging window includes layers
both before and after the intervention onset, so the plotted value
should be interpreted as a descriptive late-layer summary.
}
\label{fig:attention_dynamics}
\end{figure}

The previous subsection suggested that the prefix-conditioned prediction at
an individual token position can become less sensitive to further attention-side
intervention at later depths.
However, that criterion alone does not describe how the internal interaction
structure changes as the intervention onset is varied.
To obtain a compact descriptive view of this dependence, we examine the mean
attention entropy averaged over a fixed late-layer window as a function of
the intervention onset layer $k$.
In Fig.~\ref{fig:attention_dynamics},
we compute this quantity for representative
tokens such as \texttt{recognition} and \texttt{symbolic}, using each token
as the query position.

\paragraph{Intervention-onset dependence of mean attention entropy.}

Figure~\ref{fig:attention_dynamics} shows the intervention-onset dependence of the mean attention
entropy. 
For each intervention onset layer $k$, we compute the mean attention entropy over the fixed set of recorded layers $\ell=12,\ldots,23$: \begin{equation} \overline{H}_{\mathrm{attn}}(k) = \frac{1}{12} \sum_{\ell=12}^{23} H_{\mathrm{attn}}^{(\ell)}(k), \end{equation} where $H_{\mathrm{attn}}^{(\ell)}(k)$ denotes the attention entropy averaged over heads at layer $\ell$ in the intervention run with onset layer $k$. The same set of layers is used for every value of $k$.
For $k\leq 12$, all recorded layers lie at or after the intervention onset. For $k>12$, however, the fixed averaging range includes both layers before and layers at or after the intervention onset. The resulting quantity should therefore be interpreted as a fixed-window summary of late-layer attention behavior rather than as an average over intervention-affected downstream layers alone.
This provides a compact summary of how attention concentration changes
under different depths of context-side attention-output suppression. Lower
entropy indicates that the target token concentrates attention on a smaller
subset of positions. Importantly, the resulting curve need not vary
monotonically with the intervention onset layer. Instead, its changes provide
a coarse summary of how attention concentration responds to suppressing the
attention-side contribution from different depths.

\paragraph{Indication of regime-like behavior.}

The intervention-onset profile is clearly non-uniform. In particular,
for ``recognition,'' the fixed-window late-layer entropy summary changes
sharply between onset layers $k=2$ and $k=3$, after which it varies within
a substantially lower range. This observation indicates a marked dependence
of late-layer attention concentration on the intervention onset, but it does
not by itself identify a layer-localized transition in the ordinary forward
trajectory. When considered together with the output-side intervention-response
results, the profile is consistent with, but does not independently establish,
a regime-like organization of the depth-wise dynamics.

\paragraph{Relation to basin-like selection.}
When considered together with the intervention-response results of Section~\ref{sec:basin_completion},
the fixed-window entropy profiles are consistent with the possibility that changes
in attention concentration accompany basin-like selection. However, the present
statistic does not directly measure the number of viable candidate regions or identify
the layer at which region selection occurs. We therefore treat the entropy profile
as a descriptive companion to the output-side intervention analysis.

\paragraph{Caveat: descriptive rather than causal.}
At the same time, attention entropy should not be overinterpreted as a direct
causal mechanism. It is an observable summary statistic of the current
attention distribution, not a direct identifier of the underlying state-space
transitions. A change in $H_{\mathrm{attn}}$ may reflect the consequences of
trajectory-level reorganization rather than the cause of it. Accordingly, we
use attention entropy only as a descriptive companion to the intervention-based
analysis, not as a stand-alone proof of the proposed regime structure.

\subsection{Readout Sensitivity of the Output Token and Softmax-Level Observations}
\label{sec:output_readout}

The previous subsections focused mainly on internal sequence-state dynamics:
reduced output sensitivity under intervention and descriptive changes in
attention structure. These should be distinguished from the final readout
mechanism of the next-token softmax itself. To make that distinction explicit,
we now examine the output token from the readout side using softmax-level
quantities. 
In this subsection, we compare two component-wise interventions applied at
all token positions from layer $k$ onward. ATTN-stop suppresses the attention-output
contribution, whereas MLP-stop suppresses the MLP contribution.

For Prompt 1 and Prompt 2, we take the output token (the last token position)
as the observation point and, for interventions beginning at layer $k$, measure
relative to the baseline run (i) the rank of the baseline top-$1$ token within
the patched output distribution, and (ii) the entropy of that patched output
distribution. The former measures how well the original top candidate is
preserved under intervention, whereas the latter measures how sharp or
ambiguous the resulting softmax distribution is.

\begin{figure}[htb]
\centering
\includegraphics[width=0.7\linewidth]{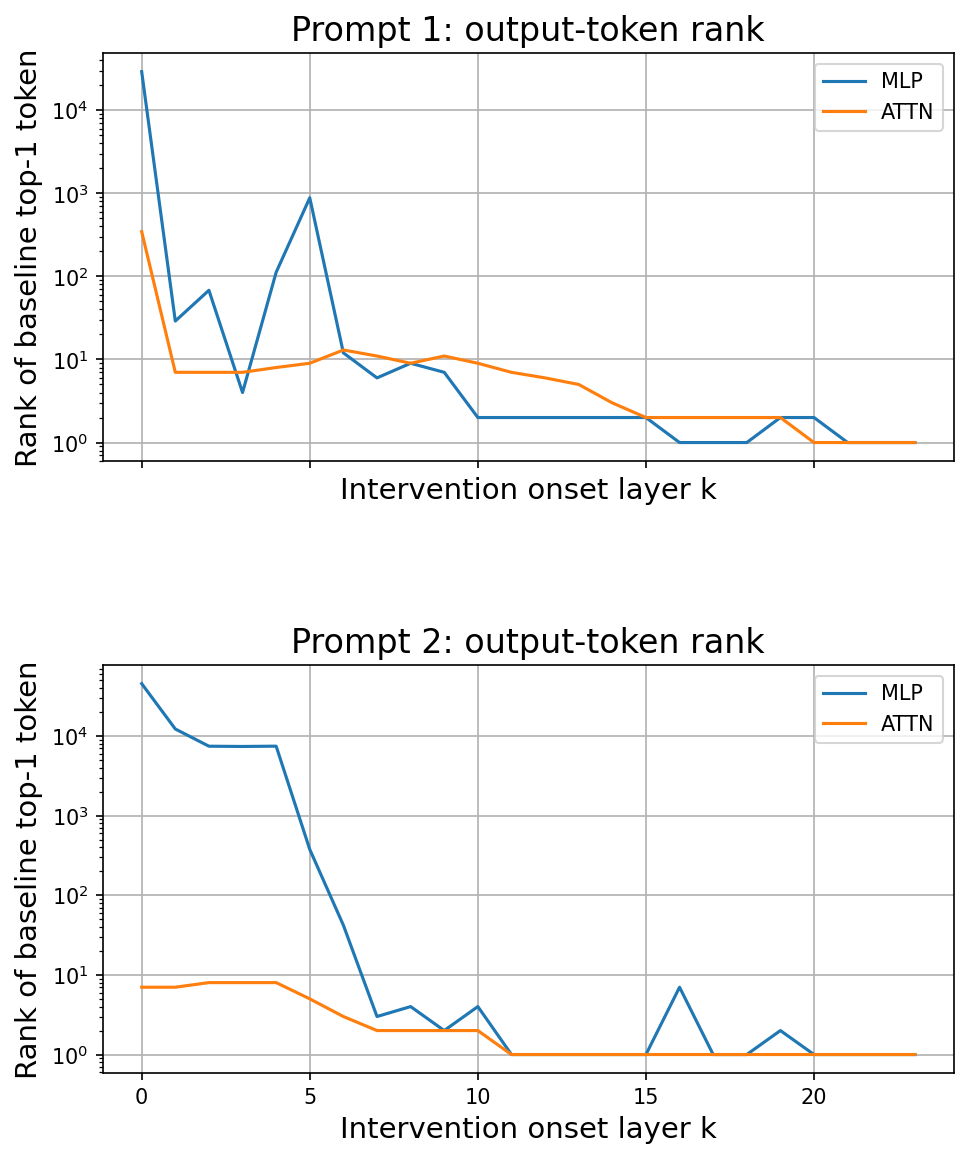}
\caption{
Readout sensitivity of the output token under MLP-stop and ATTN-stop as a
function of the intervention onset layer $k$, shown as the rank of the baseline
top-$1$ token within the patched distribution (log scale).
Top: Prompt 1; bottom: Prompt 2.
Under MLP-stop, the rank deteriorates by several orders of magnitude for
interventions beginning in shallower layers and then gradually recovers in
deeper layers. Under ATTN-stop, by contrast, the rank change is comparatively
small and tends to recover toward the baseline rank in deeper layers. This
asymmetry suggests that, for this model and intervention setting, final
vocabulary selection at the output token depends more directly on the MLP
contribution than on the attention-side contribution.
}
\label{fig:output_rank_cut}
\end{figure}

\begin{figure}[htb]
\centering
\includegraphics[width=0.7\linewidth]{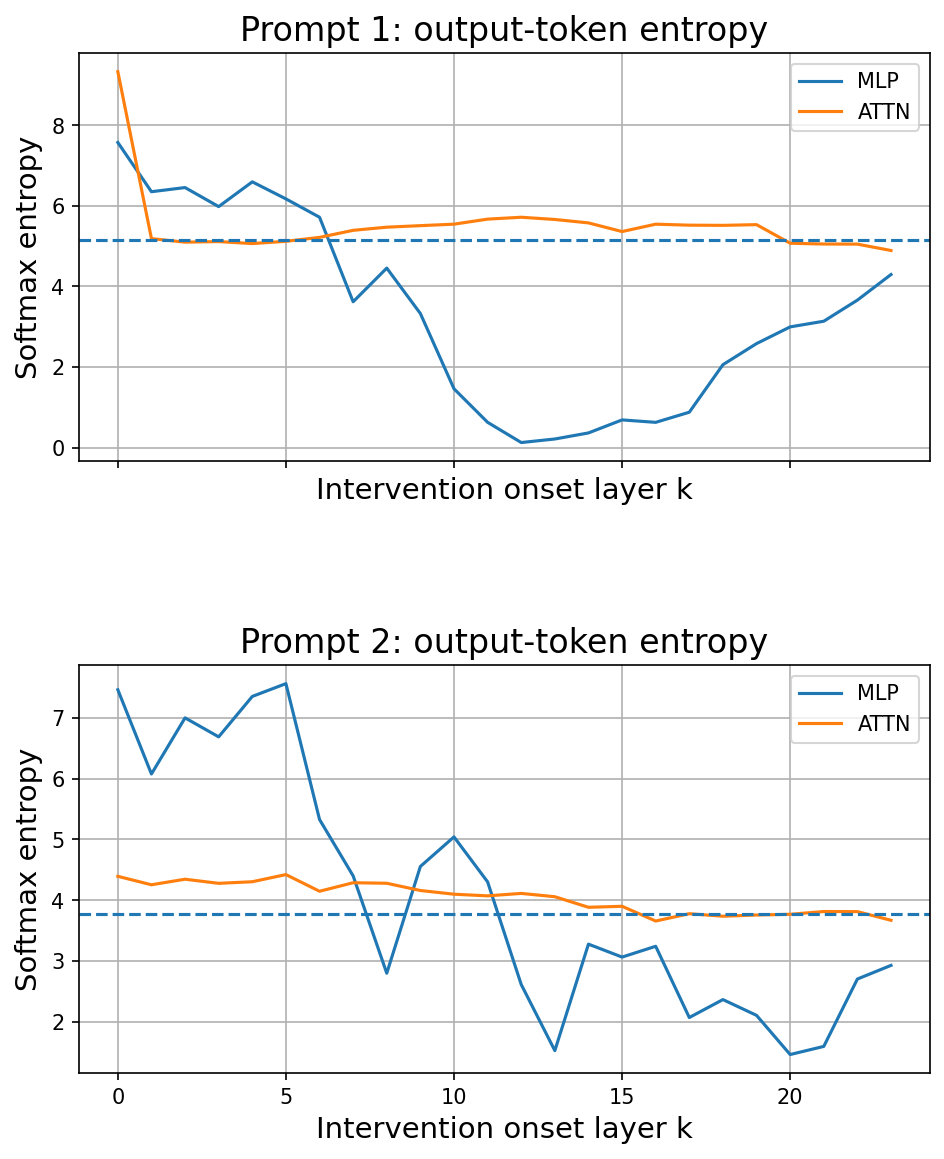}
\caption{
Softmax entropy of the patched output distribution under MLP-stop and ATTN-stop
as a function of the intervention onset layer $k$. The dashed line indicates
the baseline entropy.
Top: Prompt 1; bottom: Prompt 2.
For \(k \ge 1\),
the entropy under ATTN-stop remains relatively close to the baseline, whereas
under MLP-stop it changes substantially with layer position: in shallow layers
the distribution often becomes more diffuse, while in some middle-to-deep
layers it becomes excessively concentrated. This suggests that the MLP is
strongly involved not only in vocabulary ranking but also in controlling the
sharpness of the output distribution. These readout-side observations are
auxiliary to the main C-stop analysis and should be interpreted as
model-specific empirical evidence rather than as a claim of architectural
universality.
}
\label{fig:output_entropy_cut}
\end{figure}

Figure~\ref{fig:output_rank_cut} shows that, for both prompts, MLP-stop causes
much larger deterioration in the rank of the baseline top-$1$ token than
ATTN-stop, especially when the intervention begins in relatively shallow
layers. By contrast, under ATTN-stop, the baseline top candidate is
comparatively robust and often recovers in the deeper layers. This asymmetry
suggests that, at least for this model and this type of intervention, final
vocabulary selection at the output token depends more directly on the MLP
contribution.

Figure~\ref{fig:output_entropy_cut} supports the same picture from a different
indicator. Under ATTN-stop, the entropy of the output distribution remains
relatively stable, whereas under MLP-stop it varies strongly with depth.
Depending on the intervention depth, MLP suppression can produce either
excessive dispersion or excessive concentration of the distribution. This
indicates that the MLP is involved not only in preserving the winning vocabulary
item, but also in shaping the competition structure among candidate tokens.

These observations complement the sequence-state dynamics studied in
Sections~\ref{sec:basin_completion} and \ref{sec:entropy_transition}. A
coherent interpretation is that context-side attention-output contributions are
more important for sequence-level reorganization and basin-like selection,
whereas the MLP plays a more direct role in the final stabilization and
sharpening of the output distribution. At the same time, we emphasize that this
readout-side result is obtained for a specific GPT-style model (Pythia-410M)
and should be regarded as an auxiliary empirical observation rather than as a
claim of architectural universality.

Selected numerical values for the same ATTN-stop and MLP-stop intervention settings used in
Figs.~\ref{fig:output_rank_cut} and~\ref{fig:output_entropy_cut} are listed
in Appendix~\ref{sec:appendix_readout_values} (Table~\ref{tab:appendix_output_prompt1_rank}).

\section{Interpretation of Non-uniform Transformer Dynamics}
\label{sec:interpretation}

In this section, we integrate the observations of Section~5 and interpret Transformer inference as a structured, non-uniform sequence-level dynamical process. We focus in particular on (1) asymmetry across tokens, (2) asynchronous stabilization, and (3) the coexistence of model-dependent structure and input-dependent variation. 
In what follows, we mainly use the term in-basin stabilization, rather than convergence, to emphasize the operational and finite-horizon nature of the proposed description.

\subsection{Non-uniform and Asymmetric Dynamics}
\label{sec:nonuniform_dynamics}

The results of Sections~\ref{sec:basin_completion}--\ref{sec:output_readout} show that the depth-wise dynamics of the sequence state are non-uniform across token components. Even within the same layer, there can coexist
\begin{itemize}
  \item components that have already stabilized sufficiently with respect to the output,
  \item components that continue to evolve and still affect the readout.
\end{itemize}
This means that the residual update should not be interpreted as a single synchronized process in which all degrees of freedom evolve on the same time scale. Rather, Transformer inference behaves as an asynchronous dynamical system whose components are functionally differentiated. The observed non-uniformity is therefore not merely noise, but part of the internal organization of the computation.

\subsection{Layer-Wise Regimes as Emergent Structure}
\label{sec:layer_regime}

The combination of output-side intervention-response results and the
non-uniform dependence of the fixed-window attention-entropy summary on
intervention onset is consistent with a regime-like organization of
depth-wise Transformer inference. As an interpretive summary rather
than a directly identified set of layer boundaries, this organization may be described in terms of:
\begin{itemize}
  \item \textbf{Early stage (exploration):}
  broad interaction and competition among multiple candidate regions;
  \item \textbf{Intermediate stage (selection):}
  progressive commitment to a specific output-associated basin;
  \item \textbf{Late stage (refinement):}
  continued in-basin stabilization, with reduced sensitivity of the output to further attention-side intervention.
\end{itemize}
This staged description is compatible with conventional layer-wise summaries, but the present analysis suggests that the underlying process is both more heterogeneous and more asynchronous than a purely layer-averaged view might imply. Thus, layer-wise regimes are best understood as coarse-grained emergent descriptors of richer component-level dynamics.

\subsection{Separation of Roles Between Attention and MLP}
\label{sec:role_separation}

Within the sequence-level formulation, the experimental update takes the form
\[
H^{(l+1)}=H^{(l)}+A^{(l)}\!\left(H^{(l)}\right)+M^{(l)}\!\left(H^{(l)}\right).
\]
The observations of Section~5 support interpreting these two contributions as having distinct dynamical roles:
\begin{itemize}
  \item the Self-Attention term primarily supports sequence-level reorganization and basin selection,
  \item the MLP term primarily supports token-wise local refinement and output stabilization within the selected region.
\end{itemize}

In particular, reduced output sensitivity under later attention-side suppression indicates
that the context-side attention-output contribution becomes less consequential at later
intervention onset depths for many tokens under the present intervention criterion.
By contrast, the continued sensitivity of output-level indicators to MLP suppression suggests that MLP contributions remain important in the later stages of the trajectory. This asymmetry supports the view that attention and the MLP should not be regarded as dynamically symmetric constituents, but as components that operate at different effective scales and with different roles.

\subsection{Input-Dependent Strength and Model-Dependent Structure}
\label{sec:input_dependence}

In our exploratory analysis of multiple prompts, including additional prompts whose results are not shown, we observed qualitatively similar patterns, although their strength and duration depended on the input.
In particular,
\begin{itemize}
  \item the sharpness of the transition and the number of layers over which it persists vary with the prompt,
  \item the stabilization rate of individual tokens also depends on the input sequence.
\end{itemize}
At the same time, the approximate layer range at which transition-like behavior appears, as well as the broad early/middle/late organization,
appeared comparatively stable across the prompts examined for the fixed model.
This suggests a useful separation:
\begin{itemize}
  \item \textbf{structure:} largely model-dependent,
  \item \textbf{strength and timing details:} input-dependent.
\end{itemize}
Under this interpretation, different inputs trace different trajectories through a common meta-structure imposed by the architecture and learned parameters.

\subsection{Toward a Dynamical-Systems View of Transformers}
\label{sec:dynamical_view}

Taken together, these observations suggest that Transformer inference is better viewed not as a mere composition of layers, but as a discrete-time dynamical process on a high-dimensional sequence-level state space with the following properties:
\begin{itemize}
  \item multiple output-associated regions may exist in the state space,
  \item Self-Attention contributes sequence-level path control and region selection,
  \item the MLP contributes token-wise local refinement within the selected region,
  \item stabilization is non-uniform across the components of the sequence state.
\end{itemize}
A particularly important implication is that the effective target of inference need not be a single point in state space. Operationally, it may instead be an output-invariant region: multiple distinct trajectories can remain compatible with the same final output. From this perspective, Transformer inference is shaped not only by \emph{which} region is reached, but also by \emph{how} the trajectory enters and evolves within that region.

\subsection{Limitations of the Intervention-Based View}
\label{sec:limitations_intervention}

The present framework is intentionally operational and therefore comes with several limitations. First, the intervention used in our experiments relies on directly suppressing a residual contribution, which may in some cases move the hidden state away from the manifold typically visited during ordinary inference. Accordingly, robustness of the final output under late attention-side suppression should not be interpreted as proving that the corresponding component is generically unnecessary in a stronger architectural sense. Rather, it shows that under the present intervention criterion, the final output can remain effectively invariant despite that suppression.

Second, the present study does not yet include a systematic robustness comparison across alternative intervention schemes such as mean-ablation, freezing-based interventions, or attention-pattern-preserving variants. Such comparisons would be valuable for determining
how strongly the observed token-wise stabilization patterns and basin-like interpretations depend on the specific intervention used.

Third,
the notions of basin, invariance region, and token-wise stabilization used here
are operational and measurement-dependent. In particular, they depend on how output
difference is quantified and on the tolerance used to judge effective invariance.
These notions should therefore be understood as experimentally grounded descriptors of inference-time behavior, rather than as canonical dynamical objects independent of metric choice.

Finally, the present results are demonstrated on a specific trained GPT-style model and should be interpreted as evidence for a useful organizing framework rather than as a universal claim about all Transformer architectures. Extending the analysis across model sizes, architectures, and intervention variants remains an important direction for future work.

The purpose of the present study is therefore not to exhaustively compare
all possible intervention families, but to establish a minimal operational
framework in which token-level output invariance can be used to organize
component-wise depth dynamics.

In particular, the present experiments do not attempt to determine a universal
numerical tolerance for output invariance; the calibration of such a tolerance
requires broader empirical validation across models, prompts, and intervention
schemes.

\section{Conclusion}
We have presented an intervention-based dynamical framework for describing Transformer
inference across depth. By decomposing each residual update into additive self-attention
and MLP contributions, the framework enables component-wise interventions and measurement
of their effects through output-side responses.

The contrasting responses to ATTN-stop and MLP-stop provide intervention-based evidence
for a functional distinction between the two components. The results are consistent with
self-attention primarily supporting sequence-level direction selection, context integration,
and reorganization, whereas the MLP primarily supports position-wise representational refinement
and output stabilization. This interpretation is not intended as an exclusive decomposition of
their functions, but as an operational model of their distinguishable intervention-response profiles.

For the context-side analysis, we used C-stop, which suppresses the attention-output contribution
only at the context positions while preserving it at the final position used for next-token readout.
Although a comparison with A-stop may provide additional information
about the functional significance of the final readout position, its effect is likely to depend
on the prompt and the dependencies among its tokens. A systematic comparison is therefore left for future work.

The prompt dependence of internal computation limits the generality of the claims that can be made, 
but it does not preclude a mathematical description of Transformer inference. Rather than proposing
a prompt-independent law for token stabilization, the present framework provides an operational and
falsifiable model in which the attention and MLP contributions can be selectively suppressed and their
effects measured. Future work should examine the robustness of the proposed interpretation across
different models, prompts, tasks, and intervention settings.

\section*{Acknowledgments}
During the development of this study, generative AI tools were used as auxiliary
tools for exploring relevant literature, improving English expression, and examining
research ideas through iterative discussion. The authors critically evaluated
and verified the resulting suggestions and information.
The formulation of the research questions, the development of the methodology,
the interpretation of the results, and the preparation of the final manuscript
remain the sole responsibility of the authors.

\bibliographystyle{IEEEtran} 
\bibliography{references}

\appendix
\onecolumn

\section{Detailed Comparison with Closely Related Approaches}

For completeness, Table~\ref{tab:comparison_prior_methods_full} provides a more detailed comparison with closely related approaches, complementing the compact summary in Table~\ref{tab:comparison_prior_methods_compact} in the main text.

\begin{landscape}
\begingroup
\scriptsize
\setlength{\tabcolsep}{3pt}
\renewcommand{\arraystretch}{1.18}

\begin{longtable}{
>{\raggedright\arraybackslash}p{2.4cm}
>{\raggedright\arraybackslash}p{3.2cm}
>{\raggedright\arraybackslash}p{3.2cm}
>{\centering\arraybackslash}p{1.7cm}
>{\raggedright\arraybackslash}p{7.4cm}
}
\caption{Detailed comparison with closely related approaches.
The novelty of the present work lies not in claiming that Transformer internals
can be probed quantitatively per se, but in organizing such probing around
intervention-based output invariance at token-level resolution, with an explicit
separation between Self-Attention and MLP residual contributions.}
\label{tab:comparison_prior_methods_full}
\\
\toprule
\textbf{Work}
&
\textbf{Primary method}
&
\textbf{Evaluation / resolution}
&
\textbf{SA/MLP}
&
\textbf{Main difference from this work}
\\
\midrule
\endfirsthead

\toprule
\textbf{Work}
&
\textbf{Primary method}
&
\textbf{Evaluation / resolution}
&
\textbf{SA/MLP}
&
\textbf{Main difference from this work}
\\
\midrule
\endhead

\midrule
\multicolumn{5}{r}{Continued on next page}
\\
\endfoot

\bottomrule
\endlastfoot

Logit Lens / Tuned Lens
\cite{nostalgebraist2020logitlens,belrose2023tunedlens}
&
Layer-wise readout of latent predictions
&
Intermediate vocabulary predictions; layer/token level
&
No
&
Quantitative readout without intervention-based output invariance, and without
an explicit separation between context-side attention mixing and later local
refinement.
\\[1.5mm]

Causal Tracing / ROME
\cite{meng2022rome}
&
Causal intervention and model editing
&
Factual recall / editable computation; prompt/token level
&
Partial
&
Demonstrates causal localization of internal computation, but not token-wise
effective depth under sustained suppression across layers.
\\[1.5mm]

Ben-Artzy \& Schwartz (2024)
\cite{benartzy2024attendfirst}
&
Freeze / randomize / skip upper-layer context-side states or attention
&
Task-level performance under late-layer intervention
&
Attention-focused
&
Closest intervention-based precedent for late-layer attention redundancy;
our work differs by characterizing token-level output stability and
token-dependent intervention-response trajectories within individual prompts.
\\[1.5mm]

He et al. (2024)
\cite{he2024notallattention}
&
Redundancy analysis and pruning of Attention / MLP / blocks
&
Performance drop under pruning; task/model level
&
Yes
&
Supports the asymmetry between Attention and MLP, but does not provide an
operational dynamical interpretation in terms of token-wise invariance and
region selection.
\\[1.5mm]

Lad et al. (2025)
\cite{lad2024stages}
&
Layer deletion / swapping and depth-wise robustness analysis
&
Top-1 prediction robustness and stage-wise interpretation
&
Indirect
&
Proposes a stage-of-inference view and depth-dependent sensitivity, but not
token-level invariance under controlled interventions on decomposed residual
components.
\\[1.5mm]

Lu et al. (2019)
\cite{lu2019mpds}
&
Multi-particle / continuous-depth dynamical formulation
&
Theoretical architectural interpretation
&
Yes
&
Provides an important dynamical-systems precursor, but
does not define intervention-based output invariance or
characterize token-wise stabilization through component-specific
intervention during inference.
\\[1.5mm]

Geshkovski et al. (2023)
\cite{geshkovski2023clusters,geshkovski2023mathperspective}
&
Interacting-particle / self-attention dynamics analysis
&
Clustering, long-time behavior, and mathematical structure
&
Attention-centered
&
Provides mathematical grounding for self-attention dynamics, but not an
empirical SA/MLP decomposition with token-wise invariance under intervention.
\\[1.5mm]

\textbf{This work}
&
Residual decomposition into SA and MLP contributions with depth-wise suppression
interventions
&
Output invariance under intervention; token/prompt level
&
\textbf{Yes}
&
Organizes prior probing and intervention lines around token-level output
invariance, separates context-side attention mixing from later local refinement,
and proposes a dynamical interpretation in terms of region selection and
subsequent stabilization.
\\

\end{longtable}
\endgroup
\end{landscape}

\section{Numerical values underlying the intervention-response figures}
\label{sec:appendix_numerical_values}

In this appendix, we list the numerical values underlying the intervention-response curves shown in the main text.
For the prefix-ending token positions used
in Figs.~1--2, Tables~\ref{tab:appendix_recognition_values} and \ref{tab:appendix_symbolic_values} report the baseline and patched scalar output diagnostics for each intervention onset layer $k$. We report the baseline logit margin $m_{\mathrm{base}}$, the patched logit margin $m_{\mathrm{patch}}$, their absolute difference $|\Delta m| = |m_{\mathrm{base}}-m_{\mathrm{patch}}|$, the baseline softmax entropy $H_{\mathrm{base}}$, the patched softmax entropy $H_{\mathrm{patch}}$, and their absolute difference $|\Delta H| = |H_{\mathrm{base}}-H_{\mathrm{patch}}|$.

These tables are intended as descriptive supplements to the plotted curves.
In the present exploratory study, the reduction and relative stabilization
of $|\Delta m|$ and $|\Delta H|$ at later intervention onset layers are used
only as qualitative guides to effective invariance; they are not presented
as a universal, tolerance-independent criterion.

\begin{longtable}{rcccccc}
\caption{ Prefix ending at the token ``recognition'': numerical values underlying the intervention-response curves in Fig.~1. The full prompt is ``While neural networks excel at pattern recognition, symbolic methods.'' }
\label{tab:appendix_recognition_values}\\
\toprule
$k$ & $m_{\mathrm{base}}$ & $m_{\mathrm{patch}}$ & $|\Delta m|$ & $H_{\mathrm{base}}$ & $H_{\mathrm{patch}}$ & $|\Delta H|$ \\
\midrule
\endfirsthead
\toprule
$k$ & $m_{\mathrm{base}}$ & $m_{\mathrm{patch}}$ & $|\Delta m|$ & $H_{\mathrm{base}}$ & $H_{\mathrm{patch}}$ & $|\Delta H|$ \\
\midrule
\endhead
\midrule
\multicolumn{7}{r}{Continued on next page} \\
\endfoot
\bottomrule
\endlastfoot
0 & 0.7276 & 0.2998 & 0.4278 & 2.0689 & 3.3404 & 1.2715 \\
1 & 0.7276 & 0.0087 & 0.7189 & 2.0689 & 9.8438 & 7.7749 \\
2 & 0.7276 & 0.0470 & 0.6806 & 2.0689 & 7.2036 & 5.1347 \\
3 & 0.7276 & 0.9963 & 0.2687 & 2.0689 & 2.9151 & 0.8462 \\
4 & 0.7276 & 0.8483 & 0.1207 & 2.0689 & 4.5912 & 2.5223 \\
5 & 0.7276 & 0.9969 & 0.2693 & 2.0689 & 4.2785 & 2.2096 \\
6 & 0.7276 & 0.2183 & 0.5093 & 2.0689 & 4.9078 & 2.8389 \\
7 & 0.7276 & 0.4753 & 0.2523 & 2.0689 & 4.5168 & 2.4479 \\
8 & 0.7276 & 0.4854 & 0.2422 & 2.0689 & 3.8840 & 1.8151 \\
9 & 0.7276 & 0.4471 & 0.2805 & 2.0689 & 3.5132 & 1.4443 \\
10 & 0.7276 & 0.2824 & 0.4452 & 2.0689 & 3.0401 & 0.9712 \\
11 & 0.7276 & 0.2192 & 0.5084 & 2.0689 & 2.9430 & 0.8741 \\
12 & 0.7276 & 0.2735 & 0.4541 & 2.0689 & 2.5999 & 0.5310 \\
13 & 0.7276 & 0.0826 & 0.6450 & 2.0689 & 2.3893 & 0.3204 \\
14 & 0.7276 & 0.6191 & 0.1085 & 2.0689 & 1.8147 & 0.2542 \\
15 & 0.7276 & 0.6974 & 0.0302 & 2.0689 & 1.7338 & 0.3351 \\
16 & 0.7276 & 1.0537 & 0.3261 & 2.0689 & 1.4428 & 0.6261 \\
17 & 0.7276 & 0.9324 & 0.2048 & 2.0689 & 1.5349 & 0.5340 \\
18 & 0.7276 & 0.9319 & 0.2043 & 2.0689 & 1.6272 & 0.4417 \\
19 & 0.7276 & 0.9074 & 0.1798 & 2.0689 & 1.6724 & 0.3965 \\
20 & 0.7276 & 0.7873 & 0.0597 & 2.0689 & 1.8053 & 0.2636 \\
21 & 0.7276 & 0.8226 & 0.0950 & 2.0689 & 1.7938 & 0.2751 \\
22 & 0.7276 & 0.7978 & 0.0702 & 2.0689 & 1.8746 & 0.1943 \\
23 & 0.7276 & 0.8298 & 0.1022 & 2.0689 & 1.8697 & 0.1992 \\
\end{longtable}

\begin{longtable}{rcccccc}
\caption{ Prefix ending at the token ``symbolic'': numerical values underlying the intervention-response curves in Fig.~2. The full prompt is ``While neural networks excel at pattern recognition, symbolic methods.'' }
\label{tab:appendix_symbolic_values}\\
\toprule
$k$ & $m_{\mathrm{base}}$ & $m_{\mathrm{patch}}$ & $|\Delta m|$ & $H_{\mathrm{base}}$ & $H_{\mathrm{patch}}$ & $|\Delta H|$ \\
\midrule
\endfirsthead
\toprule
$k$ & $m_{\mathrm{base}}$ & $m_{\mathrm{patch}}$ & $|\Delta m|$ & $H_{\mathrm{base}}$ & $H_{\mathrm{patch}}$ & $|\Delta H|$ \\
\midrule
\endhead
\midrule
\multicolumn{7}{r}{Continued on next page} \\
\endfoot
\bottomrule
\endlastfoot
0 & 0.1260 & 0.1180 & 0.0080 & 4.6632 & 8.5003 & 3.8371 \\
1 & 0.1260 & 3.6759 & 3.5499 & 4.6632 & 3.6483 & 1.0149 \\
2 & 0.1260 & 2.7728 & 2.6468 & 4.6632 & 5.0990 & 0.4358 \\
3 & 0.1260 & 2.5285 & 2.4025 & 4.6632 & 5.5292 & 0.8660 \\
4 & 0.1260 & 1.7998 & 1.6738 & 4.6632 & 6.2723 & 1.6091 \\
5 & 0.1260 & 1.7009 & 1.5749 & 4.6632 & 6.3763 & 1.7131 \\
6 & 0.1260 & 1.1803 & 1.0543 & 4.6632 & 6.5166 & 1.8534 \\
7 & 0.1260 & 1.0544 & 0.9284 & 4.6632 & 6.6683 & 2.0051 \\
8 & 0.1260 & 0.6340 & 0.5080 & 4.6632 & 7.0201 & 2.3569 \\
9 & 0.1260 & 0.2616 & 0.1356 & 4.6632 & 7.0177 & 2.3545 \\
10 & 0.1260 & 0.0206 & 0.1054 & 4.6632 & 7.0529 & 2.3897 \\
11 & 0.1260 & 0.0906 & 0.0354 & 4.6632 & 6.9586 & 2.2954 \\
12 & 0.1260 & 0.1936 & 0.0676 & 4.6632 & 7.0308 & 2.3676 \\
13 & 0.1260 & 0.0269 & 0.0991 & 4.6632 & 6.7845 & 2.1213 \\
14 & 0.1260 & 0.5586 & 0.4326 & 4.6632 & 6.4609 & 1.7977 \\
15 & 0.1260 & 0.7019 & 0.5759 & 4.6632 & 6.2594 & 1.5962 \\
16 & 0.1260 & 0.4612 & 0.3352 & 4.6632 & 6.0903 & 1.4271 \\
17 & 0.1260 & 0.3350 & 0.2090 & 4.6632 & 5.5745 & 0.9113 \\
18 & 0.1260 & 0.1162 & 0.0098 & 4.6632 & 5.1565 & 0.4933 \\
19 & 0.1260 & 0.1319 & 0.0059 & 4.6632 & 4.9808 & 0.3176 \\
20 & 0.1260 & 0.1583 & 0.0323 & 4.6632 & 4.8175 & 0.1543 \\
21 & 0.1260 & 0.2300 & 0.1040 & 4.6632 & 4.7875 & 0.1243 \\
22 & 0.1260 & 0.1457 & 0.0197 & 4.6632 & 4.8539 & 0.1907 \\
23 & 0.1260 & 0.2448 & 0.1188 & 4.6632 & 4.7690 & 0.1058 \\
\end{longtable}

\section{Supplementary readout-side values for Prompt~1}
\label{sec:appendix_readout_values}

For completeness, Table~\ref{tab:appendix_output_prompt1_rank} lists selected
readout-side quantities for the output token of Prompt~1 under the same
ATTN-stop and MLP-stop intervention settings used in Figs.~4 and~5. These values provide a
compact numerical companion to the readout-side results discussed in
Section~5.4 of the main text.

\begin{table}[t]
\centering
\scriptsize
\setlength{\tabcolsep}{4pt}
\renewcommand{\arraystretch}{1.12}
\caption{Selected output-token intervention results for Prompt~1
(``It would be a good idea to use this technique when'').
The values are reported for the same ATTN-stop and MLP-stop intervention settings used in
Figs.~4 and~5.
ATTN-stop corresponds to suppression of the attention-output contribution at all
token positions from layer $k$ onward, whereas MLP-stop suppresses the MLP
contribution from layer $k$ onward.
We report the rank of the baseline
top-1 token in the patched distribution, the binary match indicator, and the
absolute scalar-diagnostic differences.}
\label{tab:appendix_output_prompt1_rank}

\begin{tabular}{
r
rr
rr
rr
rr
}
\toprule
\(k\)
&
\multicolumn{2}{c}{ATTN-stop}
&
\multicolumn{2}{c}{ATTN-stop}
&
\multicolumn{2}{c}{MLP-stop}
&
\multicolumn{2}{c}{MLP-stop}
\\
\cmidrule(lr){2-3}
\cmidrule(lr){4-5}
\cmidrule(lr){6-7}
\cmidrule(lr){8-9}
&
rank & match
&
\(|\Delta m_A|\) & \(|\Delta H_A|\)
&
rank & match
&
\(|\Delta m_M|\) & \(|\Delta H_M|\)
\\
\midrule
0  & 346 & 0 & 0.1225 & 4.1693 & 29138 & 0 & 0.7371 & 2.4124 \\
1  & 7   & 0 & 0.1058 & 0.0282 & 29    & 0 & 0.8095 & 1.1914 \\
2  & 7   & 0 & 0.0891 & 0.0598 & 68    & 0 & 0.7496 & 1.2944 \\
4  & 8   & 0 & 0.2912 & 0.0946 & 112   & 0 & 0.8887 & 1.4361 \\
6  & 13  & 0 & 0.3991 & 0.0590 & 12    & 0 & 0.8224 & 0.5588 \\
8  & 9   & 0 & 0.0861 & 0.3118 & 9     & 0 & 0.1549 & 0.7031 \\
10 & 9   & 0 & 0.0534 & 0.3865 & 2     & 0 & 1.1378 & 3.6955 \\
12 & 6   & 0 & 0.0146 & 0.5581 & 2     & 0 & 2.8155 & 5.0288 \\
16 & 2   & 0 & 0.4251 & 0.3860 & 1     & 1 & 1.2560 & 4.5282 \\
20 & 1   & 1 & 0.3152 & 0.0868 & 2     & 0 & 0.7136 & 2.1609 \\
23 & 1   & 1 & 0.0619 & 0.2665 & 1     & 1 & 0.4922 & 0.8620 \\
\bottomrule
\end{tabular}
\end{table}

\end{document}